\pdfoutput=1

\documentclass[11pt]{article}

\usepackage[final]{acl}

\urldef{\repourl}\url{https://github.com/spidersouris/llm_masc_gen}

\usepackage{times}
\usepackage{latexsym}
\usepackage[inkscapelatex=false]{svg}

\usepackage[T1]{fontenc}

\usepackage[utf8]{inputenc}

\usepackage{microtype}

\usepackage{inconsolata}

\usepackage{graphicx}

\usepackage{amssymb}
\usepackage{booktabs}
\usepackage{makecell}
\usepackage{changepage}
\usepackage{svg}
\usepackage{amsmath}
\usepackage{tabularx,ragged2e}
\usepackage{tablefootnote}
\usepackage{enumitem}
\usepackage{float}
\usepackage{longtable}
\usepackage{caption}
\usepackage{listings}
\usepackage{tcolorbox}
\usepackage{titlesec}
\usepackage{subcaption}
\usepackage{lipsum}
\usepackage{multirow}
\usepackage{textcomp}  
\usepackage{scalerel}  
\usepackage[table]{xcolor}
\usepackage{colortbl}
\usepackage{fontawesome5}

\DeclareRobustCommand{\humanemoji}{\scalerel*{\includegraphics{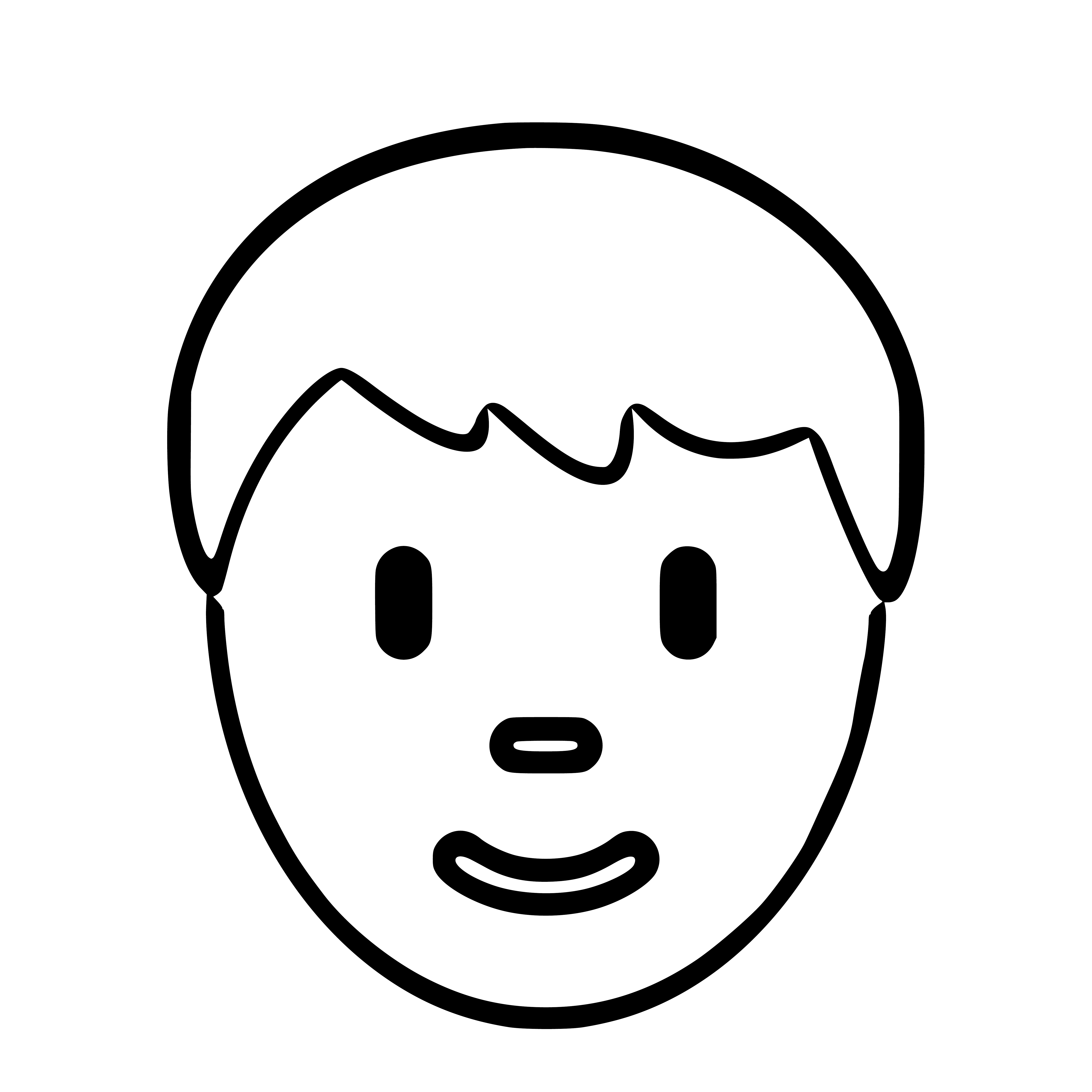}}{\textrm{\textbigcircle}}}
\DeclareRobustCommand{\robotemoji}{\scalerel*{\includegraphics{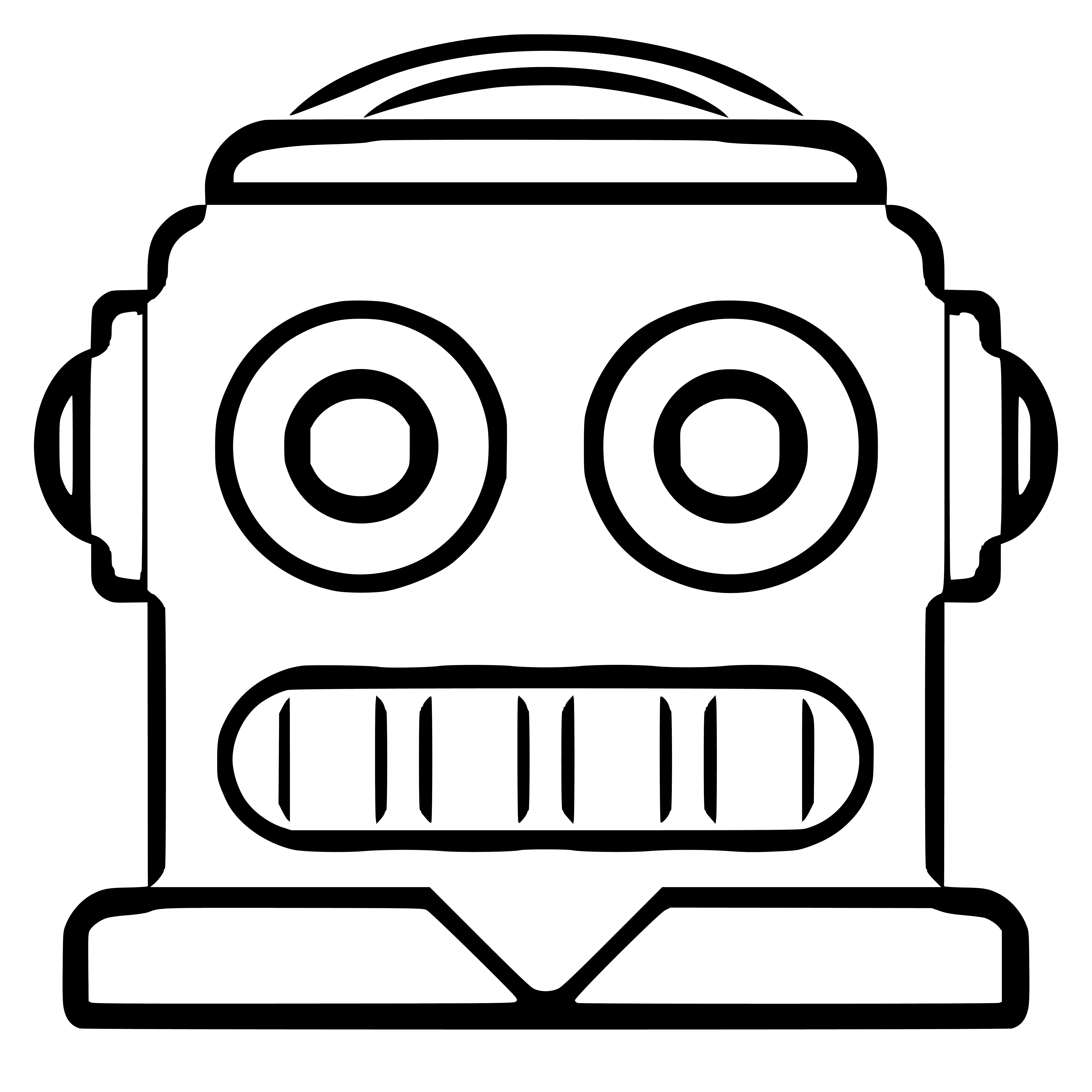}}{\textrm{\textbigcircle}}}

\definecolor{best}{RGB}{200,230,201}
\definecolor{second}{RGB}{232,245,233}
\definecolor{pareto}{RGB}{248,248,255}

\newlist{questions}{enumerate}{2}
\setlist[questions,1]{label=\textbf{RQ\arabic*.},ref=RQ\arabic*}
\setlist[questions,2]{label=(\alph*),ref=\thequestionsi(\alph*)}

\newcolumntype{L}[1]{>{\raggedright\arraybackslash}p{#1}}


\usepackage[autostyle, english = american]{csquotes}
\MakeOuterQuote{"}

\makeatletter
\newcommand{\@BIBLABEL}{\@emptybiblabel}
\newcommand{\@emptybiblabel}[1]{}
\makeatother

\usepackage[roman]{parnotes} 
\makeatletter
\def\parnoteclear{%
    \gdef\PN@text{}%
    \parnotereset }
\makeatother

\usepackage{gb4e}
\noautomath 

\makeatletter
\renewcommand{\@subex}[2]{\settowidth{\labelwidth}{#1}\itemindent\z@\labelsep#2%
         \topsep0\p@\itemsep0\p@%
         \parsep\p@\partopsep0\p@%
         \leftmargin\labelwidth%
         \ifnum\the\@xnumdepth=1
         \else\advance\leftmargin#2\relax\fi}
\makeatother

\newcommand{\mk}[1]{\texttt{#1}}
\usepackage{stfloats}

\title{Man Made Language Models? Evaluating LLMs' Perpetuation of Masculine Generics Bias}

\author{Enzo Doyen \and Amalia Todirascu \\
  LiLPa, University of Strasbourg \\
  \texttt{\{enzo.doyen,todiras\}@unistra.fr}}

\begin{document}
\maketitle
\begin{abstract}
Instruct-based large language models (LLMs) have been shown to propagate and even amplify gender bias when prompted with contextually constrained instructions (e.g., writing a text from a description or selecting a gendered pronoun). However, little attention has been paid to biases in responses to contextually unconstrained (generic) instructions conveyed by gendered language, particularly masculine generics (MG). MG, found in many gender-marked languages, denote the use of the masculine gender as a supposedly neutral reference to mixed-gender groups or individuals whose gender is unknown or non-binary. Yet, psycholinguistic studies demonstrate that MG are not neutral and systematically induce gender bias. This study investigates how both local and proprietary LLMs are MG-biased when responding to generic prompts in French, examining LLMs' MG bias rates and use of gender-fair language (GFL). We create a 16k+ human noun database from existing lexical resources and evaluate six LLMs on four instruction-response datasets under two conditions: prompts with and without MG. Overall, we find that $\approx$27.57\% of LLMs' responses to MG-filtered generic instructions are MG-biased ($\approx$78.55\% with MG-containing prompts). Moreover, we find that LLMs rarely use GFL spontaneously. These findings highlight the persistence of MG bias in LLM outputs and models’ limited tendency towards GFL strategies.
\end{abstract}

\begin{center}
    \faGithub\  \href{https://github.com/spidersouris/llm_masc_gen}{\texttt{spidersouris/llm\_masc\_gen}}
\end{center}

\section{Introduction}

Masculine generics (MG) are a linguistic feature found in many gender-marked languages, among which French, German, or Dutch. MG denote the use of the masculine gender in gendered languages as a ``default'' or supposedly neutral gender to refer to either a) a mixed group of men and women, or b) a person whose gender is irrelevant, non-binary or unknown within the context, as in the following examples in French:

\begin{exe}
\exi{(a)} \textbf{Les étudiants}\textsuperscript{MG} ont participé à l'atelier. \\ \footnotesize{(\textbf{Students}\textsuperscript{MG} participated to the workshop.)} \label{ex:1}
\
\exi{(b)} \normalsize{\textbf{Un athlète}\textsuperscript{MG} doit s'entrainer régulièrement pour progresser.} \\ \footnotesize{(\textbf{An athlete}\textsuperscript{MG} needs to train regularly to progress.)} \label{ex:2}
\end{exe}

Comparable equivalents of MG in English would translate into c) consistently using occupation nouns ending with the ``-man'' suffix or, albeit old-fashioned, d) the generic ``he'' instead of ``they'':

\begin{exe}
\exi{(c)} \textbf{The policemen}\textsuperscript{MG} started the training. \label{ex:3}
\
\exi{(d)} \underline{The reader} may skip this chapter if \textbf{he}\textsuperscript{MG} wishes. \label{ex:4}
\end{exe}

While such usage of the masculine gender supposedly acts as a neutralizer and contrasts with masculine specifics (i.e., noun phrases referring specifically to men)\footnote{For example, in the sentence ``\textbf{Cet étudiant} a participé à l'atelier.'' (``\textit{\textbf{This student} participated to the workshop.}''), the use of the masculine demonstrative determiner ``cet'' specifically indicates that the student is male.}, empirical research in psycholinguistics shows that this supposed genericness is not processed as such by native speakers \citep{gygaxGenericallyIntendedSpecifically2008,glimGenericMasculineRole2024,rothermundRemindingMayNot2024}, precisely due to the generic and the specific forms being morphologically (but not semantically) identical \citep{irmenSemanticContentGrammatical2010,gygaxMasculineFormGrammatically2021}. Psycholinguistics studies prove that MG induce cognitive biases that amplify male-centric mental representations at the expense of women \citep{braunCognitiveEffectsMasculine2005,gygaxMasculineFormIts2012,tibblinThereAreMore2023}.

The rapid development of large language models (LLMs) and their remarkable capabilities across various textual tasks have made them indispensable tools for numerous applications. Relatedly, the release of ChatGPT by OpenAI and similar chatbots through user-friendly web applications has made it extremely trivial for anyone to generate textual content. However, these models inherently reflect biases present in their training data, including those related to gender representation. Their widespread adoption introduces at least three distinct pathways through which such biases may propagate:

\par{\textbf{(1) Exposure to AI-generated content.}} As a growing amount of text on the Internet is LLM-generated \citep{brooksRiseAIgeneratedContent2024,weiUnderstandingImpactAIgenerated2024,sunAreWeAIgenerated2025,dolezalImpactAIGeneratedText2026}, people consuming this content are increasingly exposed to the related biases.
\par{\textbf{(2) LLM training on synthetic data.}} The use of synthetically generated data in the training of LLMs \citep{longLLMsDrivenSyntheticData2024,nadasSyntheticDataGeneration2025} raises the risk of recursively reinforcing existing model biases.
\par{\textbf{(3) Direct interaction with conversational LLMs.}} Many users engage with LLMs primarily for informational and decision-support purposes in conversational contexts, with \citet{chatterjiHowPeopleUse2025} reporting that 73\% of usage is non-work-related and largely interaction-based. Such chat-like exchanges may contain biased responses and could influence users' perceptions, judgments, and decisions in real time.

While previous research in Natural Language Processing (NLP; see Section \ref{sec:related}) largely discussed the issues surrounding the propagation and amplification of gender biases by LLMs in constrained contexts (e.g., instructing an LLM to continue a gender-ambiguous sentence, or to translate from/to languages with different gender systems), a crucial gap exists in the literature when it comes to the biases conveyed by LLMs' responses to contextually unconstrained (or generic) instructions due to MG use. We define generic instructions as prompts a) devoid of any reference to specific individuals and b) framed around common tasks or questions ("Practical Guidance and Seeking Information" as per \citet{chatterjiHowPeopleUse2025})\footnote{Examples of such instructions drawn from datasets used in this study are presented in Appendix \ref{apx:genericinstr}.}.




We focus on only one language with MG (French), but our methodology can be extended to any other gender-marked language given appropriate resources, that is a human noun dataset (presented in Section \ref{sec:db}), and filtered instruction/output pairs datasets (presented in Section \ref{sec:mguseinstrout}). Using these resources, we query six LLMs with filtered MG or non-MG instructions, detect MG and gender-fair language (GFL) forms (inclusive and neutral) in their responses, and validate human noun usage in context. We then evaluate the use of MG and GFL by LLMs across two experiments, drawing a comparison with human-written datasets using a novel Pareto analysis evaluation framework to take into account the multiple complementary dimensions of GFL.

\section{Related Work}
\label{sec:related}

MG and the associated male bias have been extensively studied in the field of psycholinguistics for a variety of gender-marked languages \citep{silveiraGenericMasculineWords1980,stahlbergNameYourFavorite2001,safinaEffectsGrammaticalGender2024}, including French \citep{gygaxMasculineFormIts2012,richyDemelerEffetsStereotypes2021,xiaoHowFairGenderFair2023}. For example, \citet{gygaxGenericallyIntendedSpecifically2008} tested English, French, and German speakers with role nouns (e.g., "social workers"; MG in French and German). When later asked to judge continuations referring to group members (e.g., "women"), English participants’ interpretations followed stereotypes (e.g., "beauticians" → female, "politicians" → male), while French and German participants were strongly influenced by grammatical gender: masculine plural forms consistently produced male-biased interpretations, even for female-stereotyped roles. Similarly, a survey conducted by \citet{levyLecritureInclusivePopulation2017} on the use of GFL in French showed a positive effect on mental representations. Respondents asked to name a celebrity after inclusive or gender-neutral formulations (e.g., both gendered forms, or non-gendered nouns) more often named women, whereas MG formulations prompted male-centric answers. Similar findings were reported by \citet{stahlbergNameYourFavorite2001} for German using MG role nouns.


In NLP, gender bias is by far the most studied type of bias \citep{ducelRechercheBiaisDans2024}. Studies have drawn attention to gender bias in word embeddings \citep{bolukbasiManComputerProgrammer2016}, in machine translation systems \citep{wisniewskiBiaisGenreDans2021,vanmassenhoveGenderBiasMachine2024,savoldiDecadeGenderBias2025} and in text classification tasks \citep{sobhaniFairerNLPModels2024}. More recently, LLMs too have been shown to exhibit gender biases and convey stereotypes when generating context-restricted textual content (see \citet{kotekGenderBiasStereotypes2023} for pronoun disambiguation; \citet{dollEvaluatingGenderBias2024} for pronoun prediction; \citet{youBinaryGenderLabels2024} for neutral name prediction), including in languages other than English \citep{ducelEvaluationAutomatiqueBiais2024,dingGenderBiasLarge2025}. For example, in German, \citet{schmidtTacklingGenericMasculine2026} have proposed an LLM-as-a-judge framework for measuring gender neutrality in LLM outputs, and reported that a restrictive gender-neutrality system prompt substantially increased neutral and mixed forms in LLM responses compared to a standard system prompt. Nonetheless, no studies have so far quantified MG and GFL use by LLMs in responses to generic user instructions, nor examined to what extent LLMs are prone to propagating MG-related gender bias in conversational contexts, a gap this work aims to bridge.


\section{Methodology}
\label{sec:meth}

\begin{figure*}[]
	\centering 
	\includegraphics[width=\textwidth]{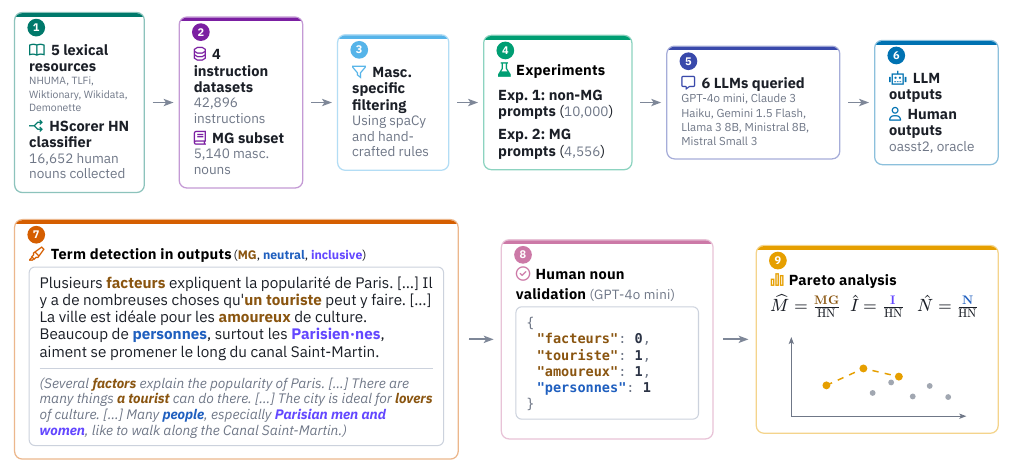}
    \caption{Overview of our methodological pipeline. We create a HN database from existing lexical resources and a custom classifier, and use a subset comprised of masculine nouns to filter out instructions referring to specific male individuals. We then divide generic instructions into two groups based on whether they contain MG or not, and query 6 LLMs. We automatically analyze LLM and human outputs for MG and GFL usage, validate human nouns with GPT-4o mini to resolve polysemic ambiguities, and compare which models and datasets are the most gender-fair using a Pareto analysis.}
	\label{fig:method}%
\end{figure*}


Our methodology is detailed in the sections below. In Section \ref{sec:db}, we create a database of French human nouns (HN) using existing online French lexical resources, and we build a French HN classifier, HScorer, to filter out nouns not referring to human beings. In Section \ref{sec:mguseinstrout}, we present LLM-generated and human-written instruction/output datasets used for our experiments and we detail our filtering process to remove masculine specific occurrences (that is, contexts where the masculine gender refers to specific male individuals and is not generic). In Section \ref{sec:analysis}, we introduce our pipeline to query LLMs with the previously filtered output datasets, automatically detect MG and inclusive terms in both LLMs' responses and human-written responses for comparison, and validate HNs in these responses with GPT-4o mini. Finally, in Section \ref{sec:measuring}, we show how we measure MG bias in human-written and LLM-inferred outputs using a novel Pareto analysis evaluation design. The complete pipeline is presented in Figure \ref{fig:method}.

\subsection{Database of French Human Nouns and HScorer}
\label{sec:db}

Since MG are nouns used strictly to refer to human beings, the first step was to create a database of French HNs to leverage for detection. To our knowledge, the only existing database in French is that of the NHUMA project \citep{stosicBaseDonneesNHUMA2013}\footnote{\url{https://nomsdhumains.weebly.com/}}, which aims to provide an extensive linguistic description of French HNs. However, this database has a limited number of entries (1,107), many of which are not MG. We used it as a basis and extended it with data scraped or collected from French lexical databases: TLFi\footnote{\url{http://atilf.atilf.fr/}}, a digitalized version of the \emph{Trésor de la langue française} dictionary; French Wiktionary\footnote{\url{https://fr.wiktionary.org/}}; and French Wikidata\footnote{\url{https://wikidata.org/}} (details in Appendix \ref{apx:hdb}).

\begin{figure}[]
	\centering 
    \includegraphics[width=0.5\textwidth]{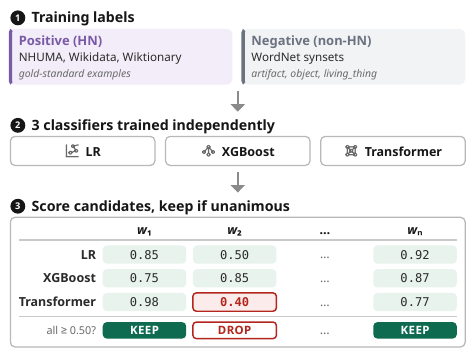}
	\caption{Overview of HScorer. We use golden HN (NHUMA, Wikidata, Wiktionary) and non-HN (WordNet) datasets to train an LR, an XGBoost and a Transformer CamemBERT model. Only words $w_n$ with full agreement as HN ($\geq$0.5) from all three classifiers are added to the database.} 
	\label{fig_hscorer}%
\end{figure}

In addition to these resources, and to further extend our dictionary, we also leveraged the Demonette database \citep{namerDemonette2DerivationalDatabase2023}, which contains pairs of masculine/feminine nouns. We also performed recursive search in TLFi to get nouns whose definition started with one of the previously TLFi-scraped nouns. However, the Demonette database includes pairs of words that do not necessarily refer to human beings (animal nouns, mythology-related nouns). Similarly, the recursive search performed on TLFi led to a number of false positives.\footnote{This could happen because HNs found during the first search may also have a non-human meaning; for example "distributeur" can be translated both to "distributor" (human) and "dispenser" (non-human).} As a potential solution to these false positives, we built an HN classifier to detect whether a noun is commonly used to refer to a human being, \textbf{HScorer}. Figure \ref{fig_hscorer} presents the classifier.

HScorer relies on three types of models: a logistic regression (LR) model, an XGBoost model \citep{chenXGBoostScalableTree2016} and a Transformer model \citep{vaswaniAttentionAllYou2017}. The LR and XGBoost model types were chosen based on their best performance compared to a set of classic machine learning model types (see Table \ref{tbl:hscorecv} in Appendix for cross-validation results). For the Transformer model, we used CamemBERT \citep{martinCamemBERTTastyFrench2020} for its best performance with French. HScorer aims at classifying nouns from the Demonette (8,048 entries)\footnote{Demonette counts exclude demonyms (21,340 entries).} and TLFi Recursive (13,961 entries) datasets as human or non-human. Only words classified as HNs by the three models (full agreement) would be part of the final HN database.

Each model was trained on golden HN and non-HN datasets. The golden HN dataset was built from previously scraped entries from Wikidata and Wiktionary, and entries from the NHUMA database. For the golden non-HN dataset, we leveraged the WordNet database \citep{princetonuniversityWordNet2010} and its collection of universal synsets through NLTK \citep{birdNLTKNaturalLanguage2004} to filter out French human-related nouns and retrieve non-HN with synsets \verb`artifact.n.01`, \verb`object.n.01` and \verb`living_thing.n.01`. We retrieved a total of 14,942 HN and 17,912 non-HN.
 
For the LR and XGBoost models, we calculated four different scores to be used as features (see Appendix \ref{apx:hscore} for feature details and Appendix \ref{apx:hscoreabl} for an ablation study). For the Transformer model, we used CamemBERT \citep{martinCamemBERTTastyFrench2020} and forwarded tokenized $\vec{W}$ as input.

We trained all three models on our golden HN and non-HN datasets with a 80/20 training-validation split using an NVIDIA RTX 4070 Super GPU. Hyperparameter optimization search was performed on all models using Weight \& Biases \citep{wandb}. Chosen hyperparameters can be found in Appendix \ref{apx:hscorehp}. An accuracy of 0.914 was achieved for the LR model; 0.937 for the XGBoost model; 0.927 for the Transformer model.

After classification, we retrieved 2,312 entries with 100\% humanness agreement from Demonette, and 1,428 from TLFi recursive search results. Our final HN database contains 16,652 unique nouns across all six initial datasets (Wiktionary, Wikidata, NHUMA, Demonette, TLFi and TLFi Recursive).

\begin{table*}[t]
\centering
\begin{tabular}{|p{1cm}|l|p{5cm}|p{1.1cm}|p{3.5cm}|p{2cm}|}
\hline
\multicolumn{1}{|c|}{\textbf{Name}} &
\multicolumn{1}{c|}{\textbf{Type}} &
\multicolumn{1}{c|}{\textbf{Description}} &
\textbf{Pairs} & \textbf{Exp. 1 \newline Filtered → Narrowed} & \textbf{Exp. 2 \newline Filtered} \\ \hline
oracle & Human & Set of question-answer pairs from French Wikipedia users & 4,613 & 2,600 → 605 & 1,338 \\ \hline
oasst2 & Human & Open Assistant Conversations Dataset Release 2 French pairs & 11,773 & 311 → 72 & 38 \\ \hline
hh\_rlhf &
LLM & French translation of Anthropic's hh-rlhf dataset, "chosen" pairs & 161,000 & 10,806 → 2,520 & 1,515 \\ \hline
alpaca &
LLM & French instruction/output pairs generated by OpenAI's GPT-3.5 & 55,184 & 29,179 → 6,803 & 1,665 \\ \hline
\multicolumn{3}{|r|}{\textbf{Total}} & \textbf{232,570} & \textbf{42,896 → 10,000} & \textbf{4,556} \\ \hline
\end{tabular}
\caption{Instruction/output datasets used for analyzing MG use. For both experiments, pairs were filtered to remove masculine specific uses (see Section \ref{sec:mguseinstrout}). Pairs for Experiment 1 were further narrowed to 10,000 total to reduce LLM querying costs.}
\label{tbl:datasets}
\end{table*}

\subsection{Instruction/Output Datasets and Masculine-Specific Filtering}
\label{sec:mguseinstrout}

Our main objective is to comprehensively evaluate the extent to which LLMs are prone to using MG in their answers to generic instructions. To this goal, we leveraged the French HN database introduced in Section \ref{sec:db} to build a MG subset (5,140 entries), that is masculine-only HNs, specifically for instruction/output filtering purposes. This subset was created by removing epicene (i.e., words whose masculine and feminine forms are identical) and feminine words from the original HN database.

Then, to analyze the use of MG by LLMs, we resorted to four LLM instruction/output datasets shown in Table \ref{tbl:datasets}: oracle\footnote{\url{https://github.com/opinionscience/InstructionFr/tree/main/wikipedia}}; oasst2\footnote{\url{https://huggingface.co/datasets/OpenAssistant/oasst2}}; hh\_rlhf\footnote{\url{https://huggingface.co/datasets/AIffl/french_hh_rlhf}}; and alpaca\footnote{\url{https://huggingface.co/datasets/jpacifico/French-Alpaca-dataset-Instruct-55K}}. The first two are human-written instructions/outputs datasets, while the last two are LLM-generated/translated. Human outputs are included to gauge the extent to which French speakers use MG in written texts, and thus have a baseline when comparing with LLM outputs.

All datasets were filtered to remove entries exhibiting specific uses of the masculine gender (i.e., contexts where a masculine word does indeed refer to a male individual), which cannot be interpreted generically and are not MG. Examples of filtered instructions and outputs are shown in Figure \ref{fig_filtering_ms}, and examples of final generic instructions are made available in Appendix \ref{apx:genericinstr}. Filtering steps are further detailed below.

\begin{figure}[]
	\centering 
	\includegraphics[width=0.50\textwidth]{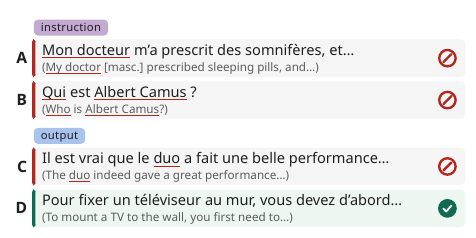}
	\caption{Examples of filtered instructions/outputs. Example A features specific masculine due to the use of a possessive determiner. Example B is filtered due to name usage. Example C contains an ambiguous term.} 
	\label{fig_filtering_ms}%
\end{figure}

Using spaCy v3.8 \citep{honnibalSpaCyIndustrialstrengthNatural2023} and the French spaCy Transformer model \verb`fr_dep_news_trf` for state-of-the-art performance, we performed POS tagging and dependency parsing on instructions from our four datasets in Table \ref{tbl:datasets}. We filtered instructions including a singular possessive determiner (e.g., "mon", 'my') or a definite determiner (e.g., "ce", 'this') followed by an HN (Ex. A in Figure \ref{fig_filtering_ms}). Similarly, instructions containing the interrogative pronoun "qui" ("who"), strictly used to refer to people, were removed as they could bias answers towards referring to a specific person; the same was applied to instructions and outputs with names (Ex. B).\footnote{Only for the "oracle" dataset, we also left out parts of instructions with "oracle" or "pythie", jargon used by the French Wikipedia community to refer to users answering inquiries.}


Finally, we specifically filtered a certain number of nouns found to be ambiguous during our preliminary tests\footnote{Full list of filtered nouns in the repository: \repourl.}. For instance, the noun "duo" in Ex. C is used to refer to several known persons in context. The list was extended by scraping a part of Wiktionary's list of 1,750 most common French words\footnote{\url{https://fr.wiktionary.org/wiki/Annexe:Liste_de_1750_mots_français_les_plus_courants}} (1,229 entries in our list in total). Moreover, as we had previously used data from Wiktionary to fill our MG dictionary, we removed all nouns from Wiktionary whose human-related definition index was greater than 2\footnote{According to the style guide of Wiktionary (\url{https://en.wiktionary.org/wiki/Wiktionary:Style_guide\#Definitions}), the most common word definitions are placed first. By retrieving only the top 2 entries, we are making sure we are only using nouns commonly used to refer to human beings.}. This helped reduce the number of nouns which were not primarily and/or inherently considered or used as HNs.

\subsection{LLM Querying and HN Validation}
\label{sec:analysis}


To gauge the use of MG by LLMs, we query a set of LLMs of different sizes and types (both proprietary and local) with the filtered instructions. Then, we automatically detect and analyze the usage of MG and gender-fair (inclusive and neutral) terms in both LLM outputs and human-written outputs for comparison. As a post-processing step, we leverage GPT-4o mini\footnote{\url{https://openai.com/index/gpt-4o-mini-advancing-cost-efficient-intelligence/}} and in-context learning \citep{brownLanguageModelsAre2020} to validate nouns considered HNs in context. This was done because, during preliminary experiments, many nouns considered MG were in fact contextually ambiguous nouns not always referring to human beings.\footnote{For instance, in French, "facteur" can refer to a mailman, but can also mean "factor, cause". Similarly, "navigateur" can refer either to a (male) sailor or to an Internet browser.}

For HN validation, GPT-4o mini was chosen for its great performance\slash cost ratio. The words and the context in which they appear were forwarded to GPT-4o mini. We used Pydantic \citep{colvinPydantic2025} to constrain its output to JSON format, each noun being a key and its value being 0 if classified as non-human, or 1 if classified as human in context. We set max tokens to 500, temperature to 0. See Appendix \ref{apx:llm-prompting} for prompting details. GPT-4o mini HN validation was evaluated by extracting 1,000 instructions from the hh\_rlhf and alpaca datasets (see Section \ref{sec:mguseinstrout}) and applying our MG use analysis pipeline. We narrow down the pipeline results to 500 to only retrieve contexts where at least one HN has been found. Those results are then sent to GPT-4o mini for validation. Two annotators classified the nouns in the 500 contexts as HN or non-HN (2,013 nouns in total). Cohen's kappa \citep{cohenCoefficientAgreementNominal1960}, noted $\kappa$, was used to compute inter-annotator agreement, where $\kappa = 0.944$. We built a reference annotation dataset and compared it with GPT-4o mini's output, where $\kappa = 0.855$. Results indicate high agreement. Divergences with GPT-4o mini are mostly due to nouns used to refer to animal or anthropomorphic characters (examples in Table \ref{tab:disgpt}).

\subsection{Measuring MG Bias}
\label{sec:measuring}

To take into account the fact that gender fairness can be expressed through multiple, partly competing linguistic strategies (inclusive markers, neutral markers and reduced use of MG), we model MG bias as a three-dimensional multiobjective optimization problem, drawing on the theory of Pareto optimality \citep{pareto1906manuale}. Each objective is associated with a distinct linguistic strategy for mitigating MG bias:

$$
\hat{M} = \frac{\mathsf{MG}}{\mathsf{HN}}, \qquad
\hat{I} = \frac{\mathsf{I}}{\mathsf{HN}}, \qquad
\hat{N} = \frac{\mathsf{N}}{\mathsf{HN}}
$$

where $\mathsf{MG}$ is the total number of MG across all texts; $I$ includes noun and pronoun inclusive forms (e.g., "il ou elle", 'he or she'; or "danseur·euses" to mean "male and female dancers") as well as neopronouns such as "iel" or "celleux" (equivalents of "they" or "zir" in English); $\mathsf{N}$ the number of neutral nouns, which gender does not depend upon the referent's (e.g., "personne", 'person'); and $\mathsf{HN}$ the total number of human nouns. Lower values of $\hat{M}$ indicate less reliance on MG, while higher values of $\hat{I}$ and $\hat{N}$ reflect greater use of inclusive and neutral GFL markers, respectively. These markers are detected using regular expressions (see Appendix \ref{apx:languagemarkers} for details). All ratios are reported in percent.

We say that a model $A$ Pareto-dominates a model $B$ if $\hat{M}_A \leq \hat{M}_B,\quad \hat{I}_A \geq \hat{I}_B,\quad \hat{N}_A \geq \hat{N}_B$ with at least one strict inequality. A model is Pareto optimal (PO) if no other model in the comparison set dominates it. We also measure MG bias by computing MG use rate in responses and comparing with GFL markers use.

\section{Experiments}

We design two experiments. In Exp. 1, we prompt LLMs with instructions containing no MG (42,896 instructions narrowed to 10,000 dataset-stratified to reduce computing and API call costs), and we compare outputs with human-written, masculine-specific-filtered responses from datasets oasst2 and oracle. In Exp. 2, we prompt LLMs with instructions containing MG only (4,556 instructions). Count details are in Table \ref{tbl:datasets}. Filtering steps explained in Section \ref{sec:mguseinstrout} were applied to only keep generic outputs. Table in Appendix \ref{apx:llmcount} shows the number of responses after filtering for both experiments. Then, HNs in responses were validated using GPT-4o mini (Section \ref{sec:analysis}). Finally, all LLM outputs had their MG bias measured (Section \ref{sec:measuring}). Steps were also applied to human-written outputs from datasets oasst2 and oracle for comparison.

For both experiments, 6 LLMs were evaluated (3 proprietary models: GPT-4o mini \citep{openaiGPT4oMiniAdvancing2024}, Claude 3 Haiku \citep{anthropicClaude3Model2024} and Gemini 1.5 Flash \citep{geminiteamGemini15Unlocking2024}; 3 local: Llama 3 8B Instruct \citep{touvronLLaMAOpenEfficient2023}, Ministral 8B \citep{mistralaiMinistralMinistraux2024} and Mistral Small 3 \citep{mistralaiMistralSmall32025}) using OpenRouter\footnote{\url{https://openrouter.ai/}}. Models were chosen to evaluate an equal amount of proprietary and local models, and based on availability through the OpenRouter API and the cost/performance ratio.

\section{Results}
\label{sec:res}

\begin{figure*}[]
	\centering 
	\includegraphics[width=0.85\textwidth]{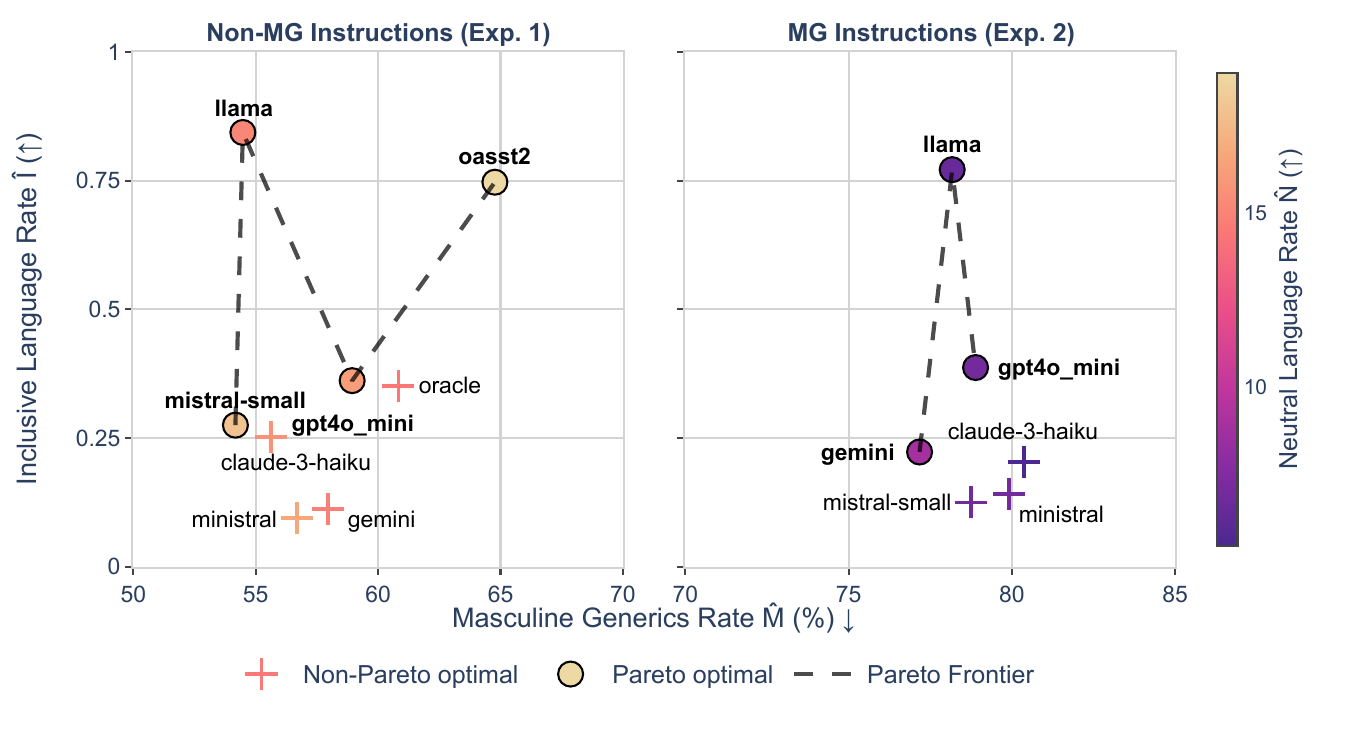}
    \caption{Pareto Frontier visualization for Experiments 1 and 2} 
	\label{fig_pareto}%
\end{figure*}

\begin{table}[t]
\centering
\footnotesize
\resizebox{\columnwidth}{!}{%
\begin{tabular}{l@{\hspace{8pt}}rccrc}
\toprule
\textbf{Model/Dataset} & {\textbf{$\hat{M}$} $\downarrow$} & {\textbf{$\hat{I}$} $\uparrow$} & {\textbf{$\hat{N}$} $\uparrow$} & \textbf{PO?} & \textbf{P(PO) \textuparrow} \\
\midrule
\multicolumn{6}{c}{\textbf{Experiment 1}} \\
\midrule
\robotemoji  claude-3-haiku  & 55.63 & 0.25 & 15.69 &  & 0.38 \\
\robotemoji  gemini          & 57.97 & 0.11 & 14.86 &  & 0.01 \\
\rowcolor{pareto}
\robotemoji  gpt4o\_mini     & 58.94 & 0.36 & 16.22 & $\checkmark$ & 0.55 \\
\rowcolor{pareto}
\robotemoji  llama           & \cellcolor{second}\underline{54.47} & \cellcolor{best}\textbf{0.84} & 15.21 & $\checkmark$ & 1.00 \\
\robotemoji  ministral       & 56.69 & 0.09 & 16.74 &  & 0.23 \\
\rowcolor{pareto}
\robotemoji  mistral-small   & \cellcolor{best}\textbf{54.17} & 0.28 & \cellcolor{second}\underline{17.99} & $\checkmark$ & 0.98 \\
\rowcolor{pareto}
\humanemoji  oasst2          & 64.77 & \cellcolor{second}\underline{0.75} & \cellcolor{best}\textbf{19.02} & $\checkmark$ & 0.96 \\
\humanemoji  oracle          & 60.83 & 0.35 & 14.51 &  & 0.14 \\
\midrule
\multicolumn{6}{c}{\textbf{Experiment 2}} \\
\midrule
\robotemoji  claude-3-haiku  & 80.37 & 0.20 & 5.45 &  & 0.00 \\
\rowcolor{pareto}
\robotemoji  gemini          & \cellcolor{best}\textbf{77.17} & 0.22 & \cellcolor{best}\textbf{8.91} & $\checkmark$ & 1.00 \\
\rowcolor{pareto}
\robotemoji  gpt4o\_mini     & 78.89 & \cellcolor{second}\underline{0.39} & \cellcolor{second}\underline{6.95} & $\checkmark$ & 0.79 \\
\rowcolor{pareto}
\robotemoji  llama           & \cellcolor{second}\underline{78.17} & \cellcolor{best}\textbf{0.77} & 6.56 & $\checkmark$ & 1.00 \\
\robotemoji  ministral       & 79.92 & 0.14 & 6.94 &  & 0.05 \\
\robotemoji  mistral-small   & 78.74 & 0.12 & 6.86 &  & 0.05 \\
\bottomrule
\end{tabular}%
}
\caption{Pareto results. Checked models are Pareto optimal. P(PO) is 95\% certainty on 10,000 resamples. Best values are bolded, second-best are underlined.}
\label{tab:pareto}
\end{table}

Table \ref{tab:pareto} reports the Pareto analysis results for both experiments, visualized in Figure \ref{fig_pareto}. In Exp. 1 (non-MG prompts), 4 models are Pareto optimal: gpt4o\_mini, llama, mistral-small, and oasst2. llama and oasst2 achieve the highest $\hat{I}$, while mistral-small has the highest $\hat{N}$. Bootstrap analysis on 10,000 resamples shows high robustness for llama, mistral-small, and oasst2 (P(PO) $\geq$ 0.96), whereas gpt4o\_mini is on the Pareto frontier less consistently (P(PO) = 0.55). In Exp. 2 (MG prompts), 3 models are Pareto optimal: gemini, gpt4o\_mini, and llama. llama displays once again the highest $\hat{I}$, while gemini achieves the lowest $\hat{M}$. All three models show high robustness under resampling, with P(PO) values ranging from 0.79 to 1.00. No single model dominates across all three dimensions in either experiment.

\begin{figure*}[]
	\centering 
	\includegraphics[width=1\textwidth]{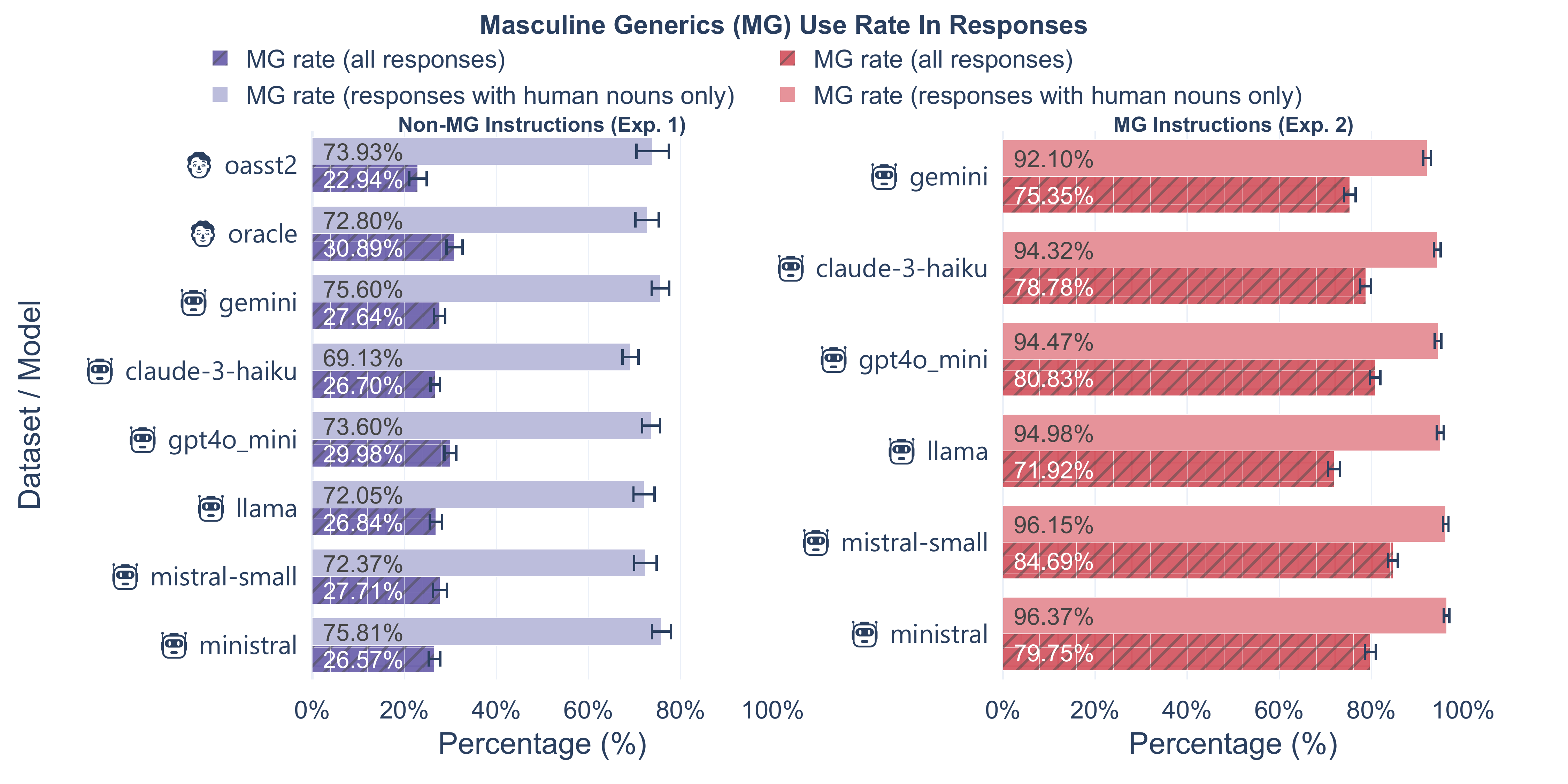}
    \caption{Percentage of MG use rate by dataset/model with $\mathsf{CI}_{95\%}$ for Experiments 1 and 2. See Appendix \ref{apx:ci} for CI details.}
	\label{fig_mguserate}%
\end{figure*}

Figure \ref{fig_mguserate} shows MG usage rates across models and datasets. On average, LLMs use MG in 27.57\% of responses to non-MG prompts (Exp. 1), rising to 78.55\% when prompts contain MG (Exp. 2). In Exp. 1, oasst2 shows significantly lower MG use rate than all LLMs based on non-overlapping confidence intervals (CIs), while differences among LLMs are not statistically reliable due to overlapping intervals. Restricting to responses with HNs only substantially increases MG rates to approximately 69-76\% across all datasets, with overlapping CIs indicating no significant differences between models or human datasets in this condition. llama has the largest gap between overall MG rate and MG rate in responses with HNs (23.06 pp).

\begin{figure*}[]
	\centering 
	\includegraphics[width=1\textwidth]{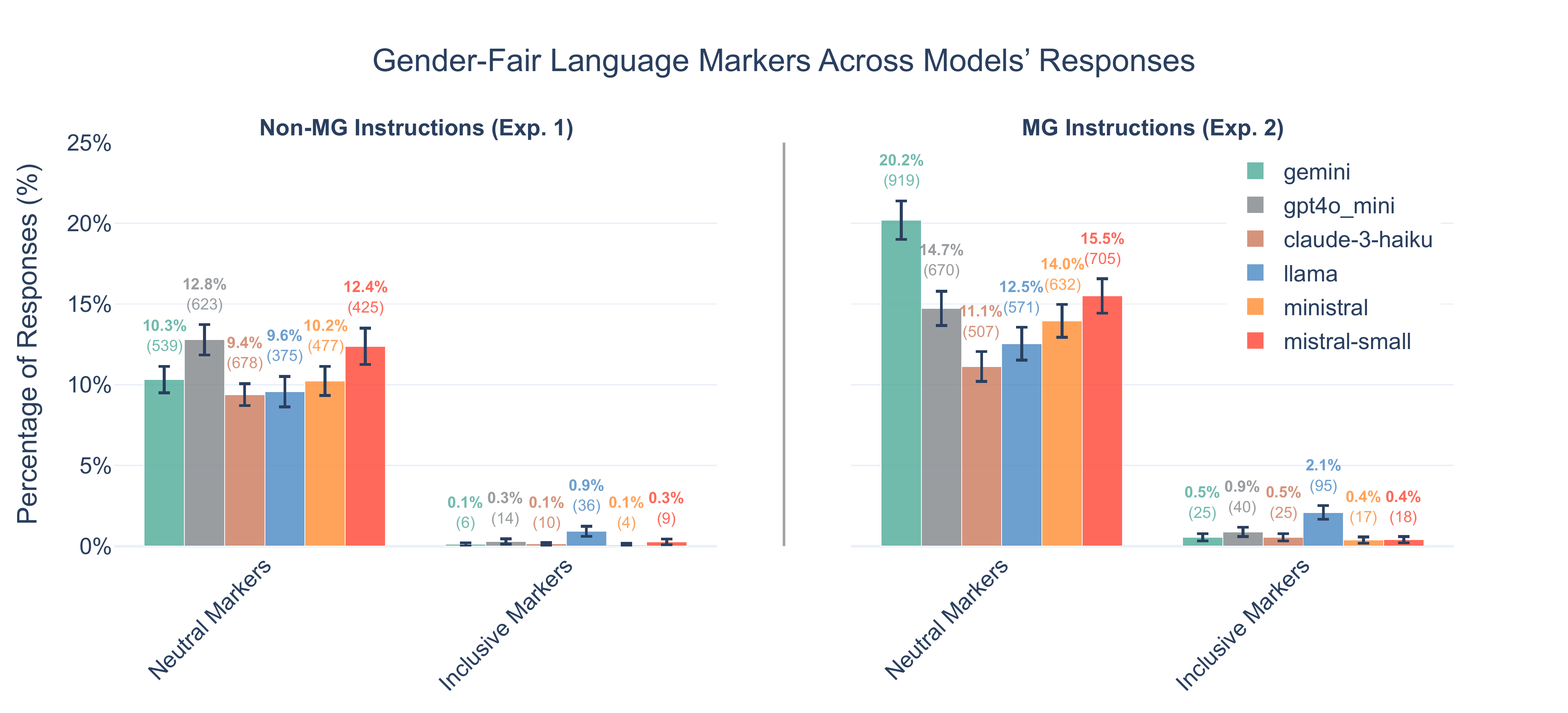}
	\caption{Percentage of responses with GFL markers across models with $\mathsf{CI}_{95\%}$ for Experiments 1 and 2. See Appendix \ref{apx:ci} for CI details.}
	\label{fig_markers}%
\end{figure*}

Figure \ref{fig_markers} indicates that LLMs are reluctant to using inclusive GFL markers spontaneously (except llama, significantly). Neutral markers are a favored option, although we see no significant difference between models (except gemini in Exp. 2, significantly).

\label{sec:disc}
\section{Discussion}

Comparing LLM and human-written responses (Exp. 1) shows that human-written texts are not necessarily less MG-biased than LLM outputs. Indeed, while oasst2 is PO, it still has the highest $\hat{M}$ compared to all models, and oracle is not on the frontier. In contrast, llama appears as more gender-fair than human-written datasets and, more generally, as the most gender-fair LLM due to its high $\hat{I}$ and low $\hat{M}$, confirmed in both experiments.

Still, bias rate analysis shows that human-written dataset oasst2 has significantly lower overall MG bias than all LLMs. In contrast, many LLMs show overall bias rates comparable to or even lower than the oracle dataset, although not always significantly given overlapping CIs. The high MG use rate in oracle may come from the fact that the dataset contains answers to questions from Wikipedia users, mostly referring to content from published articles where contextual references may inflate MG counts. Thus, while all LLMs consistently exceed the least biased human baseline (oasst2), they perform comparably to oracle, indicating that LLM bias falls between the two human datasets rather than uniformly exceeding all human-written texts. Other, larger human datasets should be used for further comparison. Similarly, when considering responses with HNs only, most models exceed or match the rates of human datasets.

MG in prompts (Exp. 2) significantly increase the prevalence of MG in LLM outputs, all while reducing GFL. While such an increase is expected for autoregressive models, it also shows the negative impact on other markers, and raises concerns about the downstream propagation of MG bias through human–LLM interaction when humans resort to MG. This highlights how users also need to be made aware of the consequences of using MG.

An analysis of GFL markers in both experiments highlights that inclusive markers are especially uncommon across all models and human-written datasets, with largely overlapping CIs (Table \ref{tbl:markersci} in Appendix \ref{apx:ci}). llama stands out due to its more frequent and statistically significant use of compact feminine endings\footnote{For example, for MG noun "vendeur" ("seller"), the feminine suffix "euse" (from "vendeuse") can be appended enclosed in parentheses in a compact form, as such: "vendeur(euse)".}, though a qualitative analysis on output examples shows that these are sometimes used incorrectly or applied to non-HN (Table \ref{tab:llmgfwrong} in Appendix \ref{apx:examplesgfl} for details). Regarding neopronouns, we find that no model uses them outside of explicit explanation contexts (e.g., claude-3-haiku in Table \ref{tab:llmgf}). This likely reflects their limited presence in training data, but it also means that LLMs do not naturally produce forms that go beyond the male–female binary. Neutral strategies, on the other hand are more common, mainly through the use of the metonymic noun ``personne'' ('person').

Additional analyses have been conducted on the types of HN used by LLMs in their responses, most particularly with regard to noun class (i.e., whether the noun refers to an occupation, a title…) and to usage specificity by LLMs. When it comes to noun class, most used HNs are occupation nouns (details in Appendix \ref{apx:defnouns}). Relatedly, we also list the top 15 most common MG nouns used by all LLMs across both experiments in Appendix \ref{apx:mgcounts}. For usage specificity, we compute a noun-specific $z$-score to find outliers in MG usage by LLMs, and find out that some nouns are much more used by some LLMs, such as ``entrepreneurs'' by llama or ``assistant'' by ministral (details in Appendix \ref{apx:outliers}).

Overall, our results indicate that LLMs consistently exhibit MG bias when generating responses to generic instructions, whether MG are part of the prompt. While previous research has clearly demonstrated that LLMs display gender bias in specific or constrained contexts, this work provides evidence that MG-induced bias can be found in everyday, instruction-based interactions with LLMs. The prevalence of MG is concerning, as it reinforces existing gender bias, strengthens male mental representations and defies efforts towards inclusion and equality. These results show that fairness in language should be attentively considered when training LLMs in languages featuring MG.

Gender bias related to the use of MG may be reduced by rewriting MG-biased texts used for training through counterfactual data augmentation \citep{luGenderBiasNeural2020}, a task known as "gender rewriting". For example, \citet{sunTheyThemTheirs2021} and \citet{vanmassenhoveNeuTralRewriterRuleBased2021} have created resources to rewrite masculine gendered pronouns in English to their gender-neutral equivalents (they/them). Similar approaches have been developed and extended to other parts of speech for Portuguese \citep{velosoRewritingApproachGender2023}, French \citep{doyenGeNReFrenchGenderNeutral2025,lernerINCLUREDatasetToolkit2024}, and Italian \citep{grecoAIAssistedInclusiveLanguage2025, piergentiliGenderNeutralRewritingItalian2025}. More recent approaches focus on using instruct-based LLMs for this task, leveraging techniques such as Retrieval-Augmented Generation \citep{muthusamychinnanGenderInclusiveLanguage2025} or task-specific instruction tuning \citep{wroblewskaIntegratingGenderInclusivity2025}.


\section{Conclusion}

Our work expands on prior research by analyzing gender bias in everyday, instruction-based interactions with LLMs. Using a French HN database created with HScorer and evaluating MG use by LLMs compared with human datasets, we find that LLMs consistently exhibit MG bias comparably to other human datasets. When evaluated in a multi-objective space, LLMs do not systematically dominate human data, showing that MG bias manifests differently across multiple dimensions. Notably, we notice a crucial lack of GFL inclusive markers use by LLMs; only Llama 3 8B Instruct marginally shows higher use of inclusive markers compared to other models. These findings highlight the need to control MG bias through multiple dimensions, and to promote GFL in LLM training data and/or during generation.

\section*{Limitations}

Our work has several limitations. First, the French HN database that we created is not exhaustive and may have missing nouns. Similarly, even though we took care in not adding non-HNs to our database, a very small number of words not used to refer to human beings may be present in the data. Second, even though we took several steps to validate human nouns, both by pre-filtering our datasets and using an LLM for automatic verification (which we evaluated), many human nouns are polysemic, and some nouns may have incorrectly been detected as human nouns, or incorrectly left undetected. Third, detection of masculine specific uses remains a challenging task. Even though we designed several filtering steps to remove instructions and responses which may contain specific uses of the masculine so as to not bias the results, a small portion might still have been included to the data. Finally, our analysis only focuses on a small set of local and proprietary LLMs, and our comparison with human-written responses is limited by only 2 French datasets. More models and more human-written datasets should be analyzed to have a broader overview of MG bias in LLMs and a better comparison point with real usage.

\section*{Ethics Statement}

The LLM instruction/output datasets were not filtered for harmful or malicious content. Similarly, LLM responses to generic instructions were not checked or filtered for such content. In addition, a very small portion of the HN database we created contains slurs and pejorative or discriminative terms used to refer to human beings. We deliberately did not filter these nouns, as they may be used in non-prejudicial contexts or with a non-harmful meaning. Finally, results of this work might be used in a counteractive way to increase gender bias in LLMs, for instance by removing occurrences of GFL markers from the training data or by increasing the number of MG occurrences in responses to generic instructions.

\section*{Acknowledgements}

We would like to thank the reviewers for their insightful and valuable comments and feedback that greatly contributed in improving this paper.

This work of the Interdisciplinary Thematic Institute LIRIC, as part of the ITI 2021-2028 program of the University of Strasbourg, CNRS and Inserm, was supported by IdEx Unistra (ANR-10-IDEX-0002), and by SFRI-STRAT’US project (ANR-20-SFRI-0012) under the framework of the French Investments for the Future Program.


\bibliography{phd}

\clearpage

\appendix
\renewcommand\thexnumi{A.5.\arabic{xnumi}}

\section{Data Collection for Dictionary Building}
\label{apx:hdb}

For TLFi, we built a custom scraper using the Playwright Node.js library\footnote{\url{https://playwright.dev/}} to retrieve words used to refer to human beings. We restricted the search to entries having one of their definitions starting with gender-neutral or gender-specific nouns or pronouns commonly used in dictionary entries related to human beings (see Table \ref{tab:defnouns}). We also performed recursive search to get nouns whose definition started with one of the previously scraped nouns.

For Wiktionary, we used Wiktextract \citep{ylonen-2022-wiktextract}\footnote{\url{https://kaikki.org/}} and applied the same filtering rule with all but one noun ("individu", 'individual')\footnote{This noun was removed from the filtering list as we had a certain number of false positives, as it can be used as a synonym of "species" and refer to living beings other than humans.}.

Following the preliminary work of \citet{lernerINCLUREDatasetToolkit2024}, we also used Wikidata to retrieve masculine-feminine HN pairs with a SPARQL query (see Appendix \ref{apx:sparql}).

\begin{table}[H]
\centering
\caption{Nouns and pronouns used to perform definition search}
\label{tab:defnouns}
\begin{tabular}{lll}
\toprule
Noun & English & Websites \\
\midrule
personne & person & TLFi, Wiktionary \\
individu & individual & TLFi \\
quelqu'un & someone & TLFi, Wiktionary \\
homme & man & TLFi, Wiktionary \\
femme & woman & TLFi, Wiktionary \\
\bottomrule
\end{tabular}
\end{table}

\subsection{Wikidata SPARQL Query Used for Human Noun Database}
\label{apx:sparql}

\begin{lstlisting}[language=xml, caption={Wikidata SPARQL Query adapted from \citet{lernerINCLUREDatasetToolkit2024}}, label=lst:sparql]
SELECT 
  ?entity 
  ?entityLabel 
  ?entityDescription
  ?entity_male_label 
  ?entity_female_label
WHERE 
{
  ?entity wdt:P2521 
    ?entity_female_label.
  OPTIONAL { 
    ?entity wdt:P3321 
      ?entity_male_label.
    FILTER(
      LANG(?entity_male_label) 
      = "fr"
    ).
  }
  SERVICE wikibase:label { 
    bd:serviceParam 
      wikibase:language "fr".
  } 
  FILTER(
    LANG(?entity_female_label) 
    = "fr"
  ).
  OPTIONAL {
    ?entity schema:description 
      ?entityDescription 
    FILTER(
      LANG(?entityDescription) 
      = "fr"
    ).
  }
  # Remove entries w/ description
  # "titre de noblesse" as many
  # dups with city names, e.g.
  # "Duc de Gênes", "Duc de Teck"
  FILTER (
    ?entityDescription 
    != "titre de noblesse"@fr 
    || 
    !BOUND(?entityDescription)
  ).
}
\end{lstlisting}

\clearpage

\section{HScorer Scoring Functions}
\label{apx:hscore}

This section details the scoring functions used as features $\vec{x}$ for the LR and XGBoost models, along with the FastText \citep{bojanowskiEnrichingWordVectors2017} vector for word embedding representation. We calculate four different scores to be used to classify each word $W$: WordNet Hypernym Score ($H$), WordNet Definition Score ($D$), FastText Score ($F$) and Suffix Score ($S$). The resulting scores, together with the FastText representation $\vec{v}(W)$ of $W$, are concatenated to form the final feature vector $\vec{v}(W)$:
$$\vec{v}(W) = [H(W) \oplus D(W) \oplus F(W) \oplus S(W) \oplus v_W]$$

\paragraph{WordNet Hypernym Score.}
Let $H(W) = (h_s, n_s)$, where:
$$h_s = \frac{\sum_{p \in P} \sum_{y \in p} 1_h(y)}{|P|}$$
$$n_s = \frac{\sum_{p \in P} \sum_{y \in p} 1_n(y)}{|P|}$$
Where $P$ is the set of hypernym paths for $W$, $1_h(y)$ is the indicator function for human hypernyms, and $1_n(y)$ is the indicator function for non-human hypernyms.

\paragraph{WordNet Definition Score.}
Let $I_h$ be a set of human indicator words and $I_n$ a set of non-human indicator words (see Table \ref{tab:seeds_indicators}). These indicator words are manually defined English words related to human ("someone", "person", "who") or non-human ("object", "plant", "chemical") entities, and are used to search in WordNet definitions. Since WordNet definitions are universally in English, we use English words. Let $D(W) = (h_d, n_d)$ where:
$$h_d = \frac{\sum_{y \in Y} \sum_{i \in I_h} f(i, \text{d}(y))}{|Y|}$$
$$n_d = \frac{\sum_{y \in Y} \sum_{i \in I_n} f(i, \text{d}(y))}{|Y|}$$
Where $Y$ is the set of synsets for $W$ and $f(i,d)$ counts occurrences of indicator $i$ in definition $d$.

\paragraph{FastText Score.}
We define a set of human prototypes $P_h$ and non-human prototypes $P_n$ (see Table \ref{tab:seeds_indicators}). These prototype words are manually defined French words prototypically related to human ("personne" (\emph{person}), "homme" (\emph{man}), "femme" (\emph{woman})) or non-human ("objet" (\emph{object}), "chose" (\emph{item}), "machine" (\emph{machine})) entities, and are used to compute semantic similarity. The prototype sets are more restrictive than the indicator sets. Let $F(W) = (h_f, n_f)$ where:
$$h_f = \frac{\sum_{p \in P_h} \cos(v_W, v_p)}{|P_h|}$$
$$n_f = \frac{\sum_{p \in P_n} \cos(v_W, v_p)}{|P_n|}$$
Where $v_W$ is the 300-dimension FastText vector for word $W$ and $\cos(v_a,v_b)$ is the cosine similarity between vectors $a$ and $b$.

\paragraph{Suffix Score.}
Finally, the suffix score captures whether $W$ ends with a human noun suffix. Specifically, $S(W) = 1$ if $W$ ends with one of the suffixes $\hat{s} \in \hat{S}$ (see Table \ref{tab:suffixes}), and $S(W) = 0$ otherwise.

The four scores are then concatenated with the FastText vector $v_W$ to obtain the final feature representation $\vec{v}(W)$ used by the LR and XGBoost models.

\begin{table}[h]
\centering
\begin{tabular}{p{0.45\linewidth} p{0.45\linewidth}}
\toprule
\textbf{Human} & \textbf{Non-Human} \\
\midrule
\multicolumn{2}{c}{\textit{Prototypes}} \\
\midrule
personne, homme, femme, individu, humain, gens &
objet, chose, machine, animal, plante, substance, outil, appareil, sentiment \\
\midrule
\multicolumn{2}{c}{\textit{Indicators}} \\
\midrule
someone, somebody, person, people, individual, professional, specialist, expert, practitioner, worker, employee, occupation, profession, who, whose, personality, character, member, leader, follower, participant, parent, child, sibling, relative &
object, thing, item, device, tool, machine, equipment, apparatus, instrument, substance, material, element, chemical, plant, animal, organism, species, concept, idea, notion, theory, principle, place, location, area, region, structure, amount, quantity, measure, unit, state, condition, situation, circumstance \\
\bottomrule
\end{tabular}
\caption{Human and Non-Human Prototypes and Indicators}
\label{tab:seeds_indicators}
\end{table}

\begin{table}[h]
\centering
\begin{tabular}{ll}
\toprule
\textbf{Masculine} & \textbf{Feminine} \\
\midrule
ant & ante \\
ent & ente \\
ateur & atrice \\
ien & ienne \\
aire & aire \\
eron & eronne \\
iste & iste \\
if & ive \\
and & ande \\
in & ine \\
\O & esse \\
on & onne \\
teur & trice \\
eur & euse \\
ier & ière \\
ais & aise \\
ois & oise \\
ard & arde \\
ain & aine \\
euf & euve \\
urge & urge \\
graphe & graphe \\
naute & naute \\
anthrope & anthrope \\
crate & crate \\
mane & mane \\
pathe & pathe \\
phile & phile \\
\bottomrule
\end{tabular}
\caption{Masculine and feminine suffixes used to compute $S$.}
\label{tab:suffixes}
\end{table}


\begin{center}
\vspace*{\fill}
\section{HScorer Cross-Validation and Final Models Hyperparameters}
\label{apx:hscorehp}

\begin{table*}[h!]
\centering
\caption{HScorer 5-Fold Cross-Validation Results on Classic Machine Learning Models}
\begin{tabular}{lllll}
\toprule
\textbf{Model} & \textbf{Precision} & \textbf{Recall} & \textbf{Accuracy} & \textbf{F1} \\
\midrule
CART & 83.50 $\pm$ 0.42 & 83.66 $\pm$ 0.49 & 85.05 $\pm$ 0.27 & 83.58 $\pm$ 0.30 \\
KNN & 85.65 $\pm$ 0.49 & 88.09 $\pm$ 0.56 & 87.87 $\pm$ 0.43 & 86.85 $\pm$ 0.47 \\
RF & 93.98 $\pm$ 0.37 & 85.61 $\pm$ 0.49 & 90.96 $\pm$ 0.32 & 89.60 $\pm$ 0.38 \\
LR & 89.07 $\pm$ 0.36 & 91.23 $\pm$ 0.35 & 90.92 $\pm$ 0.19 & 90.14 $\pm$ 0.20 \\
XGBoost & 93.96 $\pm$ 0.39 & 91.85 $\pm$ 0.29 & 93.61 $\pm$ 0.22 & 92.89 $\pm$ 0.24 \\
\bottomrule
\end{tabular}
\label{tbl:hscorecv}
\end{table*}

\titleformat{\subsection}[block]{\filcenter\normalfont\bfseries}{\thesubsection}{1em}{} 

\subsection*{Logistic Regression (LR)}
\begin{align*}
\text{penalty} &= \text{l1} \\
\text{solver} &= \text{saga} \\
\text{C} &= 100
\end{align*}

\subsection*{XGBoost}
\begin{align*}
\text{booster} &= \text{gbtree} \\
\text{learning\_rate} &= 0.22394632872649503 \\
\text{max\_depth} &= 10 \\
\text{min\_child\_weight} &= 78 \\
\text{subsample} &= 1 \\
\text{colsample\_bytree} &= 1 \\
\text{n\_estimators} &= 912 \\
\text{gamma} &= 0 \\
\text{reg\_alpha} &= 0 \\
\text{reg\_lambda} &= 0 \\
\text{objective} &= \text{binary:logistic} \\
\text{early\_stopping\_rounds} &= 20 \\
\text{random\_state} &= 42 \\
\text{tree\_method} &= \text{gpu\_hist} \\
\text{n\_jobs} &= 24
\end{align*}

\subsection*{Transformer}
\begin{align*}
\text{train\_batch\_size} &= 16 \\
\text{eval\_batch\_size} &= 16 \\
\text{eval\_strategy} &= \text{epoch} \\
\text{save\_strategy} &= \text{epoch} \\
\text{num\_train\_epochs} &= 5 \\
\text{weight\_decay} &= 0.061748150962771656 \\
\text{learning\_rate} &= 8.497821083760116 \times 10^{-6}
\end{align*}
\vspace*{\fill}
\end{center}

\section{HScorer Feature Ablation Study}
\label{apx:hscoreabl}
\raggedbottom

To gauge what are the most important features to the LR and XGBoost HScorer models, we conducted an ablation study. Table \ref{tbl:ablation} shows our different experiments, testing configurations without specific features and other configurations where only one feature is used and computing precision ($P$), recall ($R$), accuracy ($acc$) and F-score ($F1$). Features (scores) are the same as specified in Appendix \ref{apx:hscore}. Table \ref{tbl:ablation2} also shows the drop in accuracy caused by the removal of specific features.

\begin{table*}[!b]
\centering
\caption{HScorer Feature Ablation Study Results (LR and XGBoost Models)}
\label{tbl:ablation}
\begin{tabular}{lrrrrrrrr}
\toprule
experiment & $P_\mathsf{LR}$ & $R_\mathsf{LR}$ & $Acc_\mathsf{LR}$ & $F1_\mathsf{LR}$ & $P_\mathsf{XGB}$ & $R_\mathsf{XGB}$ & $Acc_\mathsf{XGB}$ & $F1_\mathsf{XGB}$ \\
\midrule
all      & 0.903 & 0.906 & 0.913 & 0.904 & 0.936 & 0.924 & 0.937 & 0.930 \\
wo\_H    & 0.893 & 0.891 & 0.901 & 0.892 & 0.920 & 0.903 & 0.920 & 0.911 \\
wo\_D    & 0.901 & 0.903 & 0.911 & 0.902 & 0.933 & 0.922 & 0.934 & 0.928 \\
wo\_F    & 0.899 & 0.908 & 0.912 & 0.904 & 0.937 & 0.924 & 0.937 & 0.931 \\
wo\_S    & 0.899 & 0.908 & 0.911 & 0.903 & 0.935 & 0.925 & 0.936 & 0.930 \\
wo\_W    & 0.792 & 0.835 & 0.825 & 0.813 & 0.794 & 0.826 & 0.823 & 0.810 \\
only\_H  & 0.597 & 0.971 & 0.688 & 0.740 & 0.600 & 0.967 & 0.690 & 0.740 \\
only\_D  & 0.584 & 0.071 & 0.553 & 0.126 & 0.529 & 0.962 & 0.591 & 0.682 \\
only\_F  & 0.746 & 0.733 & 0.764 & 0.740 & 0.739 & 0.732 & 0.760 & 0.736 \\
only\_S  & 0.660 & 0.379 & 0.628 & 0.482 & 0.660 & 0.379 & 0.628 & 0.482 \\
only\_W  & 0.883 & 0.889 & 0.896 & 0.886 & 0.916 & 0.893 & 0.914 & 0.904 \\
\bottomrule
\end{tabular}
\end{table*}

\begin{table*}[!b]
\centering
\caption{HScorer Feature Importance (LR and XGBoost Models)}
\label{tbl:ablation2}
\begin{tabular}{lrrrr}
\toprule
Feature & LR Acc. Drop & LR Drop \% & XGB Acc. Drop & XGB Drop \% \\
\midrule
H & 0.011262 & 1.233950 & 0.017045 & 1.819659 \\
D & 0.001978 & 0.216775 & 0.002435 & 0.259951 \\
F & 0.000913 & 0.100050 & -0.000609 & -0.064988 \\
S & 0.001217 & 0.133400 & 0.000304 & 0.032494 \\
W & 0.087962 & 9.638152 & 0.113681 & 12.136474 \\
\bottomrule
\end{tabular}
\end{table*}

\section{Examples of Generic Instructions}
\label{apx:genericinstr}

Examples of generic instructions from the different datasets are shown in Table \ref{tab:instructions}.

\begin{table*}[!b]
\centering
\begin{tabular}{p{0.36\textwidth} p{0.48\textwidth} p{0.10\textwidth}}
\toprule
\textbf{Instruction} & \textbf{English Translation} & \textbf{Source} \\
\midrule
Comment certains médicaments provoquent-ils une prise de poids ?
& How do some medications cause weight gain?
& hh\_rlhf \\

Rédige une publicité de 50 mots pour un nouveau parfum de luxe.
& Write a 50-word advertisement for a new luxury perfume.
& alpaca \\

Les poissons rouges peuvent-ils survivre dans du sirop de menthe ?
& Can goldfish survive in mint syrup?
& oracle \\

Explique-moi les ordinateurs quantiques comme si j'avais 10 ans.
& Explain quantum computers to me as if I were 10 years old.
& oasst2 \\

\midrule

Quelles sont les conditions pour devenir \textbf{sauveteur} ?
& What are the requirements to become \textbf{a lifeguard}?
& hh\_rlhf \\

Explique le rôle du \textbf{président des États-Unis}.
& Explain the role of the \textbf{President of the United States}.
& alpaca \\

Bonjour, je souhaiterais connaître la différence entre \textbf{un décorateur d’intérieur} et \textbf{un architecte d’intérieur}. Merci d'avance.
& Hello, I would like to know the difference between \textbf{an interior decorator} and \textbf{an interior designer}. Thank you in advance.
& oracle \\

Quels sont les stéréotypes les plus répandus concernant \textbf{les français} ?
& What are the most common stereotypes about \textbf{French people}?
& oasst2 \\
\bottomrule
\end{tabular}
\caption{Examples of generic instructions used to query LLMs from all four datasets. First 4 examples are non-MG prompts (Exp. 1); last 4 examples are MG prompts (Exp. 2) with MG nouns in bold.}
\label{tab:instructions}
\end{table*}

\section{Language Markers and MG Bias Rates Confidence Intervals}
\label{apx:ci}

To quantify statistical uncertainty for estimated MG bias rates as well as for language markers by category, we estimated 95\% bootstrap CIs using 10,000 resamples for each model and experiment.

\paragraph{MG bias rates.} Estimates are shown in Tables \ref{tbl:biasrateci} (Exp. 1) and \ref{tbl:biasrateci2} (Exp. 2). While all models/datasets exhibit substantial MG bias in both experiments, CIs show that, in Exp. 1, oasst2 has significantly lower overall bias than all LLM models. Among LLMs, however, overlapping CIs indicate that differences are not statistically significant, particularly in the middle of the ranking. In Exp. 2, llama demonstrates significantly lower bias than the most biased models (mistral-small and gpt4o\_mini), while the remaining models show overlapping intervals. For bias with human nouns (HN), CIs largely overlap across all datasets and models in both experiments, suggesting similar MG use with human referents.

\begin{table*}[!b]
\centering
\caption{Bias rate estimates (in \%) with 95\% bootstrap CIs (Exp. 1)}
\label{tbl:biasrateci}
\begin{tabular}{lrlrl}
\toprule
Dataset & Overall Bias (\%) & $\mathsf{CI}_{95\%}$ & Bias with HN (\%) & $\mathsf{CI}_{95\%}$ \\
\midrule
oracle & 30.890 & [29.150, 32.622] & 72.800 & [70.167, 75.259] \\
gpt4o\_mini & 29.980 & [28.689, 31.254] & 73.600 & [71.646, 75.509] \\
mistral-small & 27.710 & [26.224, 29.225] & 72.370 & [69.901, 74.783] \\
gemini & 27.640 & [26.436, 28.886] & 75.600 & [73.681, 77.510] \\
llama & 26.840 & [25.485, 28.214] & 72.050 & [69.756, 74.323] \\
claude-3-haiku & 26.700 & [25.689, 27.711] & 69.130 & [67.379, 70.841] \\
ministral & 26.570 & [25.322, 27.811] & 75.810 & [73.756, 77.864] \\
oasst2 & 22.940 & [21.045, 24.834] & 73.930 & [70.388, 77.446] \\
\bottomrule
\end{tabular}
\end{table*}

\begin{table*}[!b]
\centering
\caption{Bias rate estimates (in \%) with 95\% bootstrap CIs (Exp. 2)}
\label{tbl:biasrateci2}
\begin{tabular}{lrlrl}
\toprule
Model & Overall Bias (\%) & $\mathsf{CI}_{95\%}$ & Bias with HN (\%) & $\mathsf{CI}_{95\%}$ \\
\midrule
mistral-small & 84.690 & [83.660, 85.727] & 96.150 & [95.551, 96.746] \\
gpt4o\_mini & 80.830 & [79.705, 81.970] & 94.470 & [93.758, 95.172] \\
ministral & 79.750 & [78.578, 80.941] & 96.370 & [95.750, 96.946] \\
claude-3-haiku & 78.780 & [77.590, 79.939] & 94.320 & [93.576, 95.056] \\
gemini & 75.350 & [74.072, 76.555] & 92.100 & [91.231, 92.948] \\
llama & 71.920 & [70.562, 73.221] & 94.980 & [94.236, 95.686] \\
\bottomrule
\end{tabular}
\end{table*}

\paragraph{Language markers.} Estimates are shown in Table \ref{tbl:markersci}. For inclusive language markers in Experiment 1, all models show very low rates (below 1\%). llama shows the highest rate at 0.92\%, with a CI of [0.64, 1.23], making it distinguishable from most other models due to no overlap. However, for other models, the CIs overlap substantially due to very few occurrences. As a result, we cannot reliably say whether differences in these models' inclusive markers rates reflect true differences or simply sampling variation. The same applies to Experiment 2, where llama once again shows a statistically significant rate at 2.09\% with a CI of [1.70, 2.51], much higher than other models.

When it comes to neutral language markers, in Experiment 1, gpt4o\_mini [11.86, 13.73] and mistral-small [11.32, 13.51] are distinguishable from the lower cluster. In Experiment 2, gemini shows a statistically significant rate at 20.20\% with a CI of [19.03, 21.38]. Otherwise, similarly to inclusive markers, other models show much overlap.

\begin{table*}[!b]
\centering
\caption{Language markers estimates (in \%) by category with 95\% bootstrap CIs (Exp. 1)}
\label{tbl:markersci}
\begin{tabular}{lrlrl}
\toprule
Dataset & Incl. (\%) & $\mathsf{CI}_{95\%}$ & Neut. (\%) & $\mathsf{CI}_{95\%}$ \\
\midrule
oracle & 0.33 & [0.14, 0.54] & 9.62 & [8.54, 10.74] \\
oasst2 & 0.31 & [0.10, 0.56] & 9.06 & [7.78, 10.34] \\
gemini & 0.11 & [0.04, 0.21] & 10.32 & [9.51, 11.14] \\
gpt4o\_mini & 0.29 & [0.14, 0.45] & 12.79 & [11.86, 13.73] \\
claude-3-haiku & 0.14 & [0.06, 0.22] & 9.39 & [8.72, 10.07] \\
llama & 0.92 & [0.64, 1.23] & 9.57 & [8.65, 10.51] \\
ministral & 0.09 & [0.02, 0.17] & 10.24 & [9.36, 11.14] \\
mistral-small & 0.26 & [0.12, 0.44] & 12.39 & [11.32, 13.51] \\
\bottomrule
\end{tabular}
\end{table*}

\begin{table*}[!b]
\centering
\caption{Language markers estimates (in \%) by category with 95\% bootstrap CIs (Exp. 2)}
\label{tbl:markersci2}
\begin{tabular}{lrlrl}
\toprule
Dataset & Incl. (\%) & $\mathsf{CI}_{95\%}$ & Neut. (\%) & $\mathsf{CI}_{95\%}$ \\
\midrule
gemini & 0.55 & [0.35, 0.77] & 20.20 & [19.03, 21.38] \\
gpt4o\_mini & 0.88 & [0.62, 1.17] & 14.74 & [13.72, 15.79] \\
claude-3-haiku & 0.55 & [0.35, 0.77] & 11.13 & [10.23, 12.05] \\
llama & 2.09 & [1.70, 2.51] & 12.59 & [11.66, 13.56] \\
ministral & 0.38 & [0.20, 0.57] & 13.96 & [12.97, 14.98] \\
mistral-small & 0.40 & [0.22, 0.59] & 15.51 & [14.47, 16.57] \\
\bottomrule
\end{tabular}
\end{table*}





\section{Number of Non-Specific LLM Responses}
\label{apx:llmcount}

\begin{table}[H]
\centering
\caption{Number of LLM responses without specific masculine markers by model}
\label{tab:mgllmresponses}
\begin{tabular}{lrr}
\toprule
\multicolumn{1}{c}{} & \multicolumn{2}{c}{\textbf{Experiments}} \\
\cmidrule(lr){2-3}
\textbf{Model} & \textbf{Experiment 1} & \textbf{Experiment 2} \\
\midrule
claude-3-haiku   & 7,221 & 4,556 \\
gemini           & 5,224 & 4,551 \\
gpt4o\_mini      & 4,873 & 4,548 \\
ministral        & 4,660 & 4,528 \\
llama            & 3,920 & 4,452 \\
mistral-small    & 3,432 & 4,547 \\
\bottomrule
\end{tabular}
\end{table}

\clearpage

\section{MG Human Noun Classes}
\label{apx:defnouns}

We measure what types of HNs are most used as MG. To that end, we leverage the NHUMA dataset (used to create our HN dictionary in Section \ref{sec:db}), which contains human-annotated HN types (789 annotated nouns with labels such as "profession", "doer", "relationship" or "status"; see Table \ref{tab:classes} for the list of all labels). We complete these annnotations using a spaCy model trained by \citet{papasseudiClassificationNomsDHumains2023} on NHUMA data for automatic HN type labelling (see Table \ref{tab:classesbarbara} for details) and we apply it to nouns in our MG dictionary missing human annotation (i.e., non-NHUMA entries). Entries not detected as HNs by the model were left unannotated. In total, our HN dataset contains 2,893 annotated entries (i.e., +2,104 after applying the model).

\begin{table}[H]
\centering
\caption{Human noun classes and counts used by \citet{papasseudiClassificationNomsDHumains2023} for spaCy model training}
\label{tab:classesbarbara}
\begin{tabular}{ccc}
\toprule
\textbf{Class} &
\textbf{Count} & \textbf{Mapped to} \\
\midrule
NH-Mét & 697 & profession \\
NH-Fonc & 402 & profession \\
NH-Spé & 385 & speciality \\
NH-Titre & 111 & title \\
NH-Grade & 16 & title \\
\bottomrule
\end{tabular}
\end{table}

\begin{table}[H]
\centering
\caption{Human noun classes and counts across entire human noun database (including non-MG)}
\label{tab:classes}
\begin{tabular}{lrr}
\toprule
\textbf{Class} & \textbf{Count} \\
\midrule
profession   & 2,893 \\
demonym      & 1,290 \\
doer         & 388   \\
speciality   & 269   \\
attribute    & 179   \\
relationship & 44    \\
status       & 24    \\
title        & 18    \\
patient      & 14    \\
other        & 7     \\
recipient    & 5     \\
\bottomrule
\end{tabular}
\end{table}

\begin{figure*}[b]
	\centering 
	\includegraphics[width=1\textwidth]{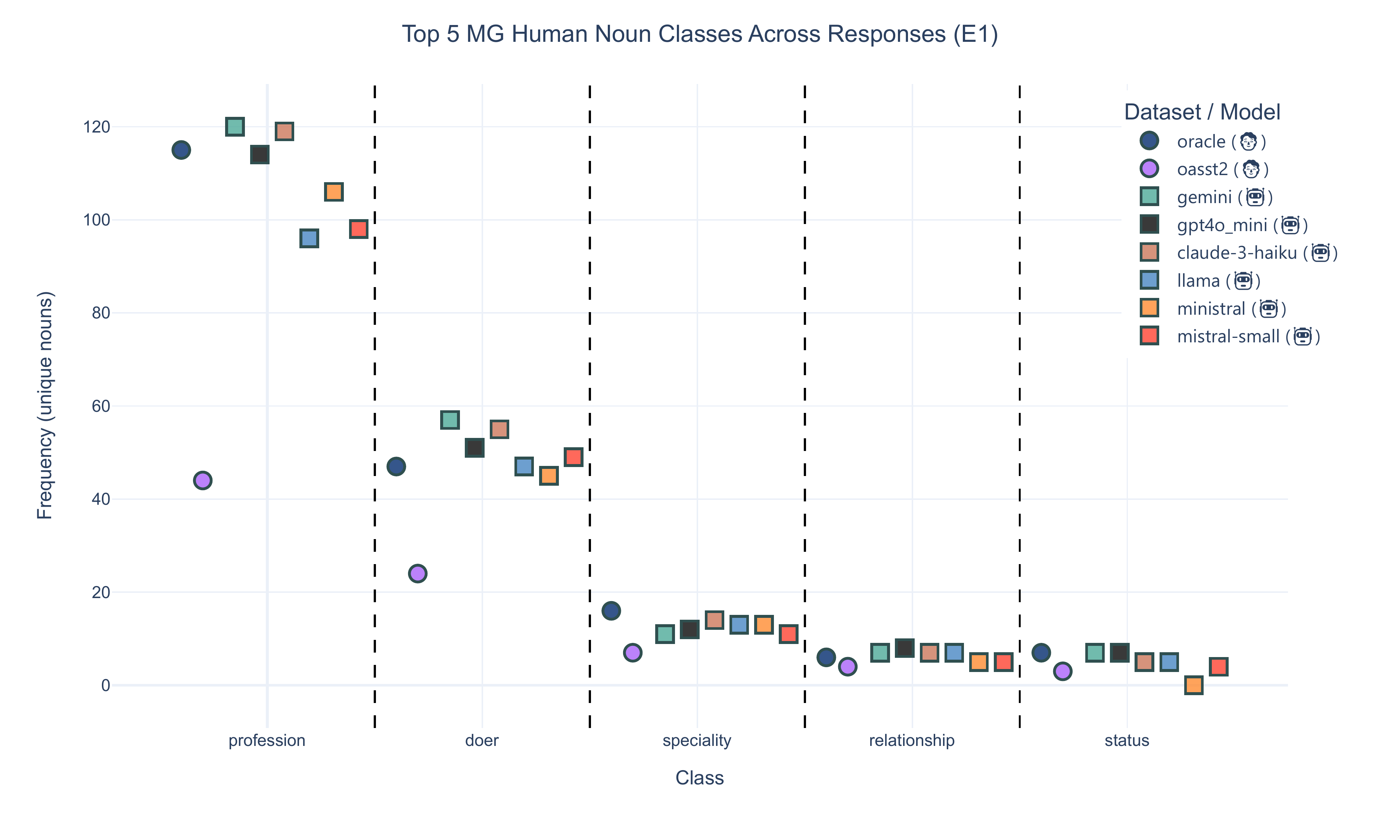}
	\caption{Frequency of unique MG nouns in responses based on their associated human noun class (Exp. 1)} 
	\label{fig_hnclasses}%
\end{figure*}

\begin{figure*}[b]
	\centering 
	\includegraphics[width=1\textwidth]{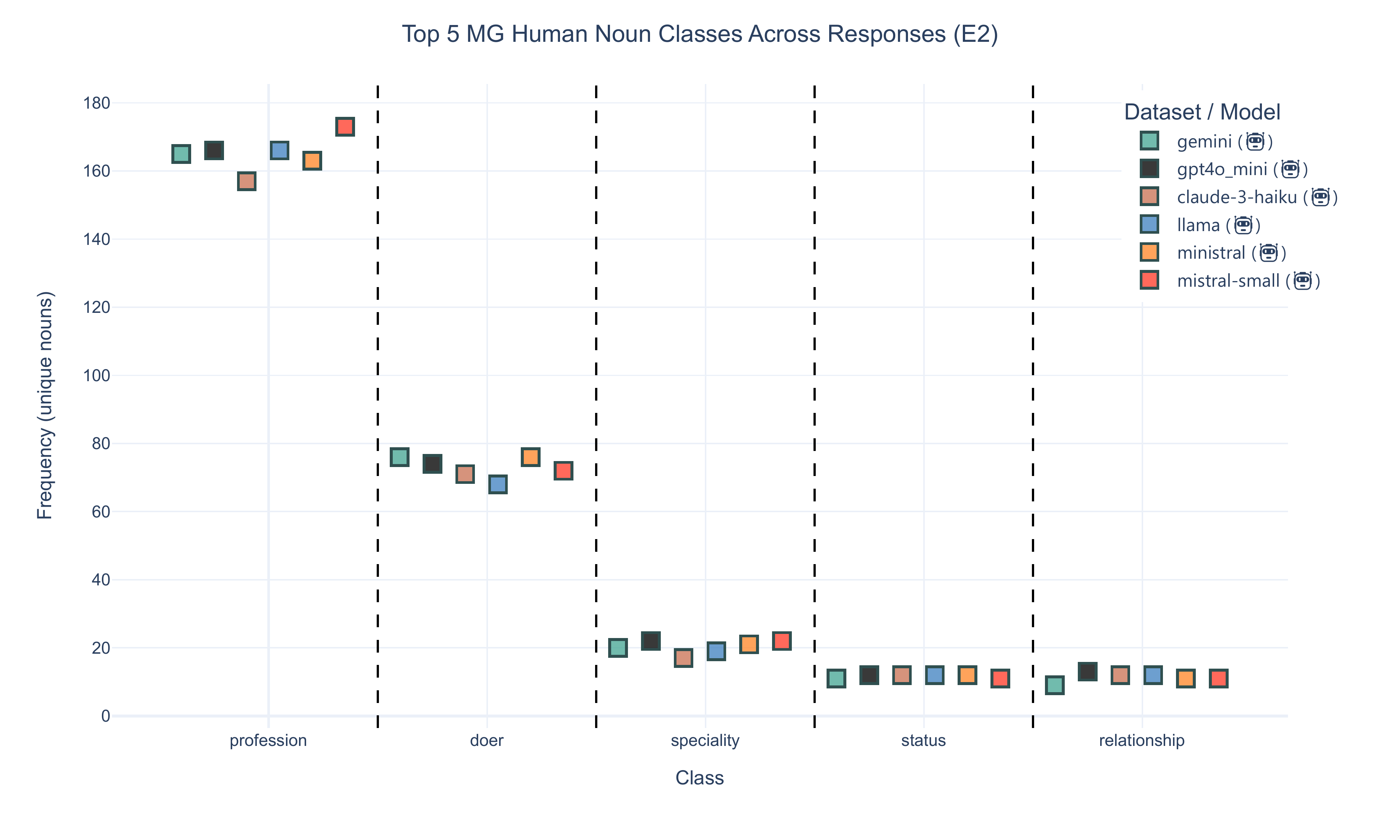}
	\caption{Frequency of unique MG nouns in responses based on their associated human noun class (Exp. 2)} 
	\label{fig_hnclasses2}%
\end{figure*}

\clearpage

\twocolumn

\section{MG Counts}
\label{apx:mgcounts}

\begin{table}[H]
\centering
\caption{Most used MG across LLM responses (Exp. 1)}
\label{tab:mostmg1}
\begin{tabular}{rlrrr}
\toprule
\textbf{Rank} & \textbf{Noun} & \textbf{English} & \textbf{Count} \\
\midrule
1 & professionnel & professional & 1,251 \\
2 & utilisateur & user & 1,057 \\
3 & ami & friend & 991 \\
4 & habitant & resident & 977 \\
5 & joueur & player & 767 \\
6 & client & customer & 590 \\
7 & consommateur & consumer & 535 \\
8 & citoyen & citizen & 469 \\
9 & homme & man & 409 \\
10 & humain & human & 380 \\
11 & auteur & author & 347 \\
12 & parent & parent & 347 \\
13 & employé & employee & 318 \\
14 & médecin & doctor & 301 \\
15 & patient & patient & 301 \\
\bottomrule
\end{tabular}
\end{table}

\begin{table}[H]
\centering
\caption{Most used MG across LLM responses (Exp. 2)}
\label{tab:mostmg2}
\begin{tabular}{rlrrr}
\toprule
\textbf{Rank} & \textbf{Noun} & \textbf{English} & \textbf{Count} \\
\midrule
1 & ami & friend & 6,497 \\
2 & homme & man & 3,589 \\
3 & étudiant & student & 2,665 \\
4 & client & customer & 2,452 \\
5 & invité & guest & 2,391 \\
6 & utilisateur & user & 2,159 \\
7 & auteur & author & 2,021 \\
8 & président & president & 1,988 \\
9 & monsieur & mister & 1,667 \\
10 & joueur & player & 1,629 \\
11 & adolescent & teenager & 1,559 \\
12 & humain & human & 1,533 \\
13 & parent & parent & 1,242 \\
14 & acteur & actor & 1,239 \\
15 & candidat & candidate & 1,148 \\
\bottomrule
\end{tabular}
\end{table}

\newpage

\section{MG Outliers with $z$-score}
\label{apx:outliers}

To determine which nouns are disproportionately frequent for certain LLMs, we identify outlier MG counts using a noun-specific $z$-score. The standard formula for $z$-score is:

$$
z = \frac{x - \mu}{\sigma}
$$

Where $x$ is the observed value, $\mu$ the mean of all observations, and $\sigma$ the standard deviation.

For each noun, we first compute the mean MG count across all observations:

$$
\mu_{\text{noun}} = \frac{1}{n}\sum_{i=1}^{n} x_i
$$

Where $x_i$ denotes the MG count of the noun in observation $i$.

We then calculate the standard deviation of the MG counts for each noun:

$$
\sigma_{\text{noun}} =
\sqrt{
\frac{1}{n-1}
\sum_{i=1}^{n}
\left(x_i-\mu_{\text{noun}}\right)^2
}
$$

The $z$-score for each observation is subsequently computed as:

$$
z_i =
\frac{x_i-\mu_{\text{noun}}}
{\sigma_{\text{noun}}},
$$

\begin{figure*}[b]
	\centering 
	\includegraphics[width=1\textwidth]{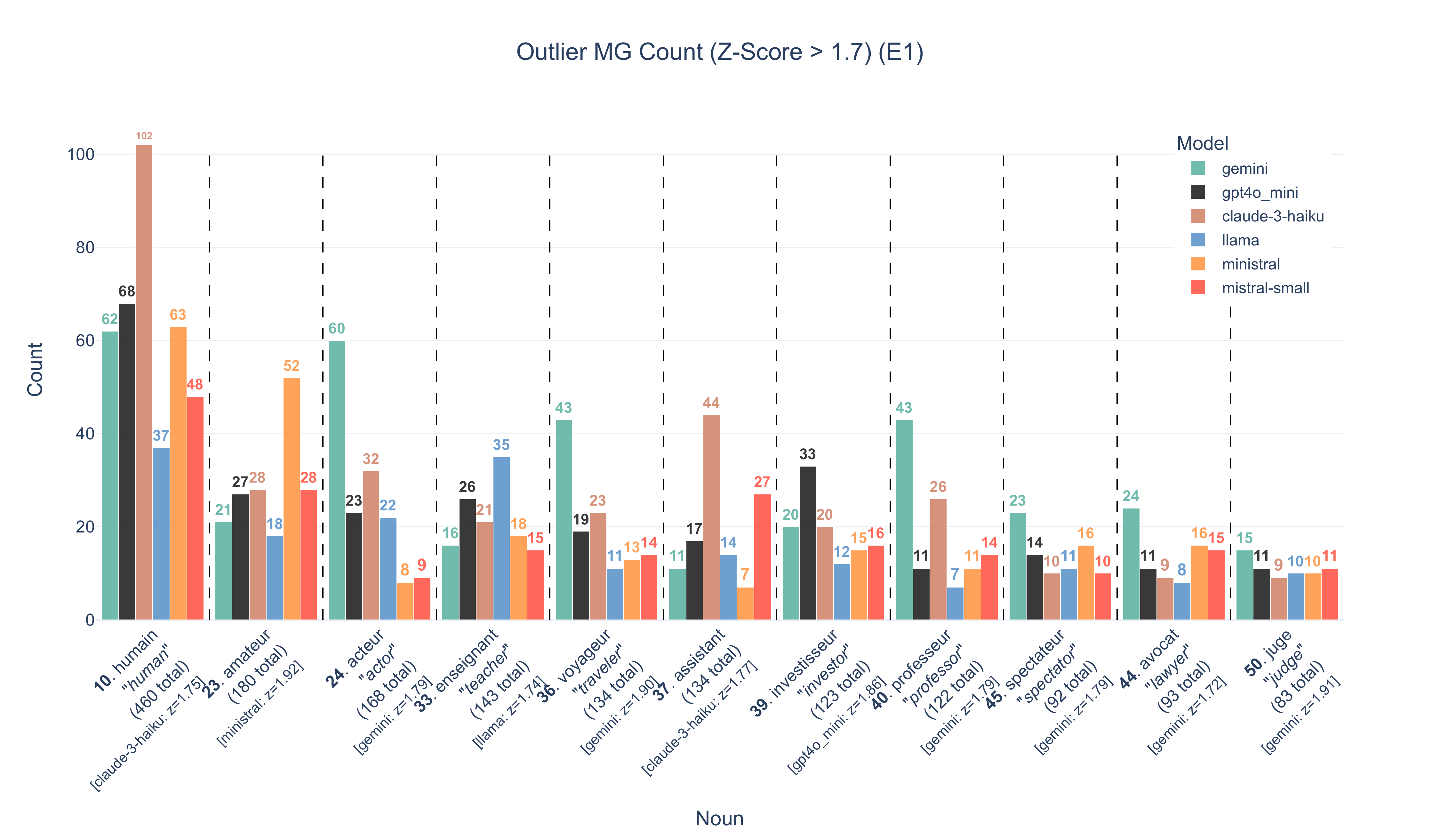}
	\caption{Counts of MG featuring model-specific outlying uses, with a $z$-score greater than 1.7 (Exp. 1)} 
	\label{fig_outliers}%
\end{figure*}

\begin{figure*}[b]
	\centering 
	\includegraphics[width=1\textwidth]{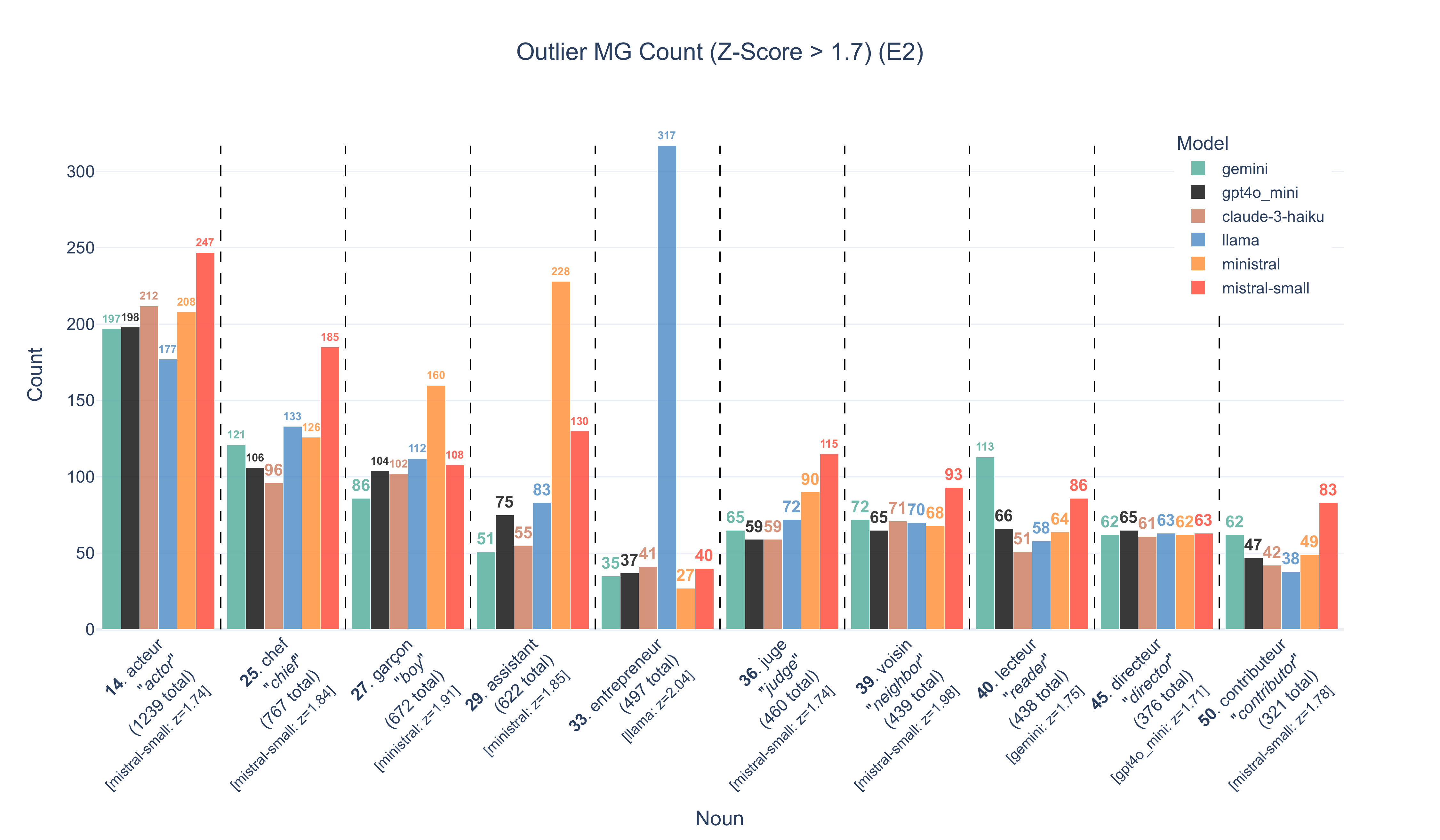}
	\caption{Counts of MG featuring model-specific outlying uses, with a $z$-score greater than 1.7 (Exp. 2)} 
	\label{fig_outliers2}%
\end{figure*}

Nouns are classified as outliers when their $z$-score exceeds $1.7$. This value was adjusted empirically. Results are displayed in Figure \ref{fig_outliers} for Experiment 1, and Figure \ref{fig_outliers2} for Experiment 2.

\clearpage

\onecolumn
\section{GPT-4o mini Human Noun Validation Prompting Details}\label{apx:llm-prompting}

\begin{center}
\begin{tcolorbox}[width=0.8\linewidth, boxrule=3pt, colback=gray!20, colframe=gray!20]
\begin{lstlisting}[caption={System Prompt}, label=lst:system-prompt, basicstyle=\rmfamily, columns=fullflexible]
You are an assistant that validates human noun classifications in French texts.
\end{lstlisting}
\end{tcolorbox}

\begin{tcolorbox}[width=0.8\linewidth, boxrule=3pt, colback=gray!20, colframe=gray!20]
\begin{lstlisting}[caption={User Prompt}, label=lst:user-prompt, basicstyle=\rmfamily, columns=fullflexible]
Given a text and nouns, for each noun, determine if it is a human noun in context. Some nouns may appear multiple times in the text. In such cases, they are distinguished by ID ('noun_1', 'noun_2'...), following the order in which they appear. Do not assume that all occurrences of the same noun are either human or non-human; instead, assess each occurrence individually based on its unique context. Only respond in this format, where human_noun is the noun being considered.
{{
  "human_noun": 0,
  "human_noun_2": 1
}}

## Examples
Text: Les facteurs d'employabilité des facteurs, chargés de distribuer le courrier, vont évoluer.
Nouns: facteurs, facteurs_2
Output: {{ "facteurs": 0, "facteurs_2": 1 }}

Text: Le président a annoncé aux citoyens une série de mesures pour renforcer l'économie du pays.
Nouns: président, citoyens, mesures
Output: {{ "président": 1, "citoyens": 1, "mesures": 0 }}

Text: Il croit aux esprits et aux fantômes depuis qu'il est enfant.
Nouns: esprits, fantômes, enfant
Output: {{ "esprits": 0, "fantômes": 0, "enfant": 1 }}

Text: {text}
Nouns: {human_nouns}
Output:
\end{lstlisting}
\end{tcolorbox}
\end{center}

\clearpage
\twocolumn

\subsection{Examples of Disagreements with GPT-4o mini}

Table \ref{tab:disgpt} displays examples of disagreements between human annotators and GPT-4o mini on the 500 contexts sampled for validation. HNs incorrectly classified as such by GPT-4o are \textbf{in bold}.

\begin{table}[h!]
\centering
\parbox{16.5cm}{\caption{Examples of disagreements between human annotators and GPT-4o mini on HN annotation}\label{tab:disgpt}}
\begin{tabular}{p{7cm}p{7cm}}
\toprule
Text & English Translation \\
\midrule
Le papillon se repose un moment pour que ses ailes se développent et sèchent, avant de pouvoir s'envoler à la recherche de nourriture et de \textbf{partenaires}. & The butterfly rests for a while so that its wings can develop and dry before it can take flight in search of food and \textbf{mates}. \\
\hline
En chemin, ils rencontrèrent de nombreux \textbf{alliés} : un sage hibou qui leur enseigna des sorts puissants, un groupe de \textbf{nains} courageux qui leur forgèrent des armes magiques, et un vieux dragon sage qui leur révéla la vérité sur l'ombre maléfique. & Along the way, they encountered many \textbf{allies}: a wise owl who taught them powerful spells, a group of courageous \textbf{dwarves} who forged magical weapons for them, and an old, wise dragon who revealed to them the truth about the malevolent shadow.  \\
\bottomrule
Il est vrai que de nombreux films anciens de Disney montrent des \textbf{personnages} fumant, ce qui peut sembler étrange aujourd'hui, étant donné les connaissances sanitaires actuelles concernant le tabagisme. & It is true that many older Disney films depict \textbf{characters} smoking, which may seem strange today, given current health knowledge about smoking.  \\
\bottomrule
\end{tabular}
\end{table}

\clearpage

\section{GFL Markers}\label{apx:languagemarkers}

Table \ref{tab:inclmarkers} displays examples of GFL markers automatically detected for the analyses. \verb|fem_ending| refers to occurrences where the feminine ending of an HN is separated from the original, masculine form to increase its visibility. This is done either by adding a special separator (e.g., "auteur·ice", "auteur(ice)"; see Tables \ref{tab:allmarkers} and \ref{tab:separators}) or by spelling the feminine ending with capitals (e.g., "auteurICE"). 

We also include, in Table \ref{tab:allmarkers}, the list of marker types included in the computation of each metric for the Pareto Frontier (Section \ref{sec:measuring}). The corresponding separators used to detect the \verb|fem_ending| type are listed in Table \ref{tab:separators}.

\begin{table}[H]
\centering
\begin{tabular}{lll}
\toprule
Example & English Translation & Type \\
\midrule
mesdames et messieurs & Ladies and gentlemen & incl\_greetings \\
tous et toutes & everyone & incl\_greetings \\
un ou une & a [m] or a [f] & incl\_pairs \\
il ou elle & he or she & incl\_pairs \\
iel & they (sg.) & neutral\_prons \\
auteur·ice; auteur(ice) & author & fem\_ending \\
\bottomrule
\end{tabular}
\caption{Examples of inclusive markers by type, with their English equivalents}
\label{tab:inclmarkers}
\end{table}

\newcommand{\hn}{\textit{HN}}
\newcommand{\exx}[1]{\textit{#1}}
\newcommand{\pat}[1]{\texttt{#1}}

\begin{table*}[h!]
\centering
\small
\setlength{\tabcolsep}{6pt}
\renewcommand{\arraystretch}{1.15}
\begin{tabularx}{\textwidth}{@{}l c >{\raggedright\arraybackslash}p{0.34\textwidth} X@{}}
\toprule
\textbf{Type} & \textbf{Metric} & \textbf{Matching pattern} & \textbf{Examples} \\
\midrule

incl\_greetings &
$I$ &
Fixed inclusive greeting expressions, plural form &
\exx{tous et toutes}, \exx{toutes et tous}, \exx{messieurs, dames},
\exx{messieurs dames}, \exx{mesdames et messieurs},
\exx{messieurs et mesdames}
\\
\addlinespace

incl\_pairs &
$I$ &
\texttt{[masculine/feminine form] (et/ou) [corresponding form]} &
\exx{ceux et celles}, \exx{celles ou ceux}, \exx{ils et elles},
\exx{il ou elle}, \exx{le ou la}, \exx{un ou une}
\\
\addlinespace

fem\_ending &
$I$ &
Inclusive language separators (see Table \ref{tab:allmarkers}) &
\pat{auteur·ices}, \pat{auteur(ice)s} …
\\
\addlinespace

neutral\_prons &
$I$ &
Gender-neutral pronouns &
\exx{iel}, \exx{iels}, \exx{ielle}, \exx{ielles},
\exx{celleux}, \exx{elleux}, \exx{ille}, \exx{illes}
\\
\addlinespace

neutral\_nouns &
$N$ &
Common neutral nouns &
\exx{gens}, \exx{personne}, \exx{personnes}, \exx{individu}, \exx{individus}
\\

\bottomrule
\end{tabularx}
\caption{List of markers by type, with corresponding metric, matching pattern and examples}
\label{tab:allmarkers}
\end{table*}
 
\begin{table}[!b]
\centering
\small
\begin{tabular}{@{}cll@{}}
\toprule
\textbf{Glyph} & \textbf{Code point} & \textbf{Unicode Name} \\
\midrule
\textperiodcentered      & U+00B7 & \textsc{middle dot} \\
\textperiodcentered      & U+2027 & \textsc{hyphenation point} \\
\ensuremath{\cdot}       & U+22C5 & \textsc{dot operator} \\
\ensuremath{\bullet}     & U+2219 & \textsc{bullet operator} \\
\textbullet              & U+2022 & \textsc{bullet} \\
\texttt{.}               & U+002E & \textsc{full stop} \\
\texttt{/}               & U+002F & \textsc{solidus} \\
\texttt{(}               & U+0028 & \textsc{left parenthesis} \\
\bottomrule
\end{tabular}
\caption{List of French inclusive language separators (placed between the stem and the feminine suffix, as in \mk{étudiant\textperiodcentered es}) for the fem\_ending type}
\label{tab:separators}
\end{table}

\clearpage

\subsection{Examples of Gender-Fair LLM Outputs}
\label{apx:examplesgfl}

We conduct a qualitative analysis of some LLM output examples from both experiments to determine how well they can use GFL in unconstrained and unprompted/open-ended contexts. Examples of incorrect GFL usage are listed in Table \ref{tab:llmgfwrong}: the wrong GFL marker is bolded and the Table includes corresponding explanations as to why these usages are wrong. Errors mostly come from an overapplication of GFL techniques to nouns that do not need it (e.g., nouns referring to objects or non-human entities), or the use of a separator suffix that does not work because the feminine equivalent of the masculine noun is morphologically very different. Conversely, Table \ref{tab:llmgf} shows examples of correct GFL usage by LLMs, with the GFL marker in bold and the corresponding passage English translation as a footnote.

\begin{table*}[h!]
\centering
\parbox{16.5cm}{\caption{Examples of incorrect gender-fair LLM outputs by model and marker type}\label{tab:llmgfwrong}}
\begin{tabular}{ccp{3.5cm}p{5cm}}
\toprule
Model (Exp.) & Marker Type & Output & Explanation \\
\midrule
gemini (E2) & incl\_pairs & **Finaliser le lieu et la date avec les complices:** Vérifier avec le gérant du lieu ou \textbf{l'hôte(lle)}. & "hôtel" (hotel) is an inanimate noun and does not change with gender, so using a gender separator is not appropriate. However, interestingly, the MG noun "gérant" (manager) used previously does not have the same separator applied, while it could have: "gérant(e)". \\
\hline
llama (E2) & incl\_pairs & Cher/e \textbf{Monsieur(e)} [Enseignant], Je vous écris pour solliciter une réunion avec vous concernant ma note obtenue & The model was asked to write a letter to a teacher. For the salutation, even though "Dear" ("cher", "chère") was correctly written separating the feminine suffix, the equivalent of "Monsieur" ("Mister") for a woman is "Madame", not "Monsieure". Applying a separator that way does not work. \\
\hline
llama (E2) & incl\_pairs & Bonjour! Je serais ravi de vous fournir une liste de vocabulaire français de base pour les débutants. Voici quelques notions essentielles pour commencer à apprendre le français [...] Soeur (sister) * \textbf{Mari(e)} (husband/wife) & The model was asked to provide a list of basic French vocabulary for a language learner. The equivalent of "mari" ("husband") for a woman is "femme", not "marie". Applying a separator that way does not work. \\
\bottomrule
\end{tabular}
\end{table*}

\clearpage

\onecolumn

\begin{table}[h!]
\centering
\parbox{13.5cm}{\caption{Examples of correct gender-fair LLM outputs by model and marker type}\label{tab:llmgf}}
\begin{tabular}{ccp{8.5cm}}
\toprule
Model (Experiment) & Marker Type & Output \\
\midrule
mistral-small (E1) & incl\_pairs & Vous soulevez un point intéressant. L'expression "c'est une égalité parfaite" est souvent utilisée dans le langage courant pour renforcer l idée de parité ou d'équilibre absolu entre deux choses. […] Par exemple, si quelqu'un dit "c'est une égalité parfaite entre nos deux équipes," \textbf{il ou elle}\textsuperscript{*} veut dire que les deux équipes sont très équilibrées en termes de force et de capacité. Pour gérer ces situations, vous pouvez soulager ce sentiment en disant que "l'expression 'de parfaite équité' ou 'd'équilibre parfait' pourrait mieux exprimer l'idée d'une égalité absolue." \\
\hline
gemini (E1) & fem\_ending & Les médias sociaux, bien que présentant des avantages en termes de connexion et d'information, engendrent également des effets négatifs. La comparaison constante avec des versions idéalisées de la vie des autres \textbf{utilisateur·ices}\textsuperscript{†} peut entraîner une baisse de l'estime de soi et de l'anxiété. Une exposition excessive à des contenus négatifs, polarisants ou manipulatoires peut affecter le bien-être mental et contribuer à la propagation de fausses informations. Enfin, la dépendance aux notifications et la distraction constante peuvent entraver la concentration et la vie sociale réelle.  \\
\hline
llama (E1) & incl\_greetings & \textbf{Mesdames et messieurs}\textsuperscript{‡}, Aujourd'hui, je voudrais vous parler d'un sujet qui concerne tous les habitants de notre planète : la sauvegarde des espèces menacées. C'est un problème immense et urgent qui nécessite notre attention et notre action. Mais avant de vous présenter les raisons pour lesquelles la sauvegarde des espèces menacées est une question essentielle, je voudrais poser une question simple : qu'adviendrait-il si nous perdions définitivement certaines espèces qui vivent sur terre? [...] \\
\hline
claude-3-haiku (E2) & neutral\_pron & Je comprends votre point de vue. La question de l'utilisation des pronoms peut être un sujet délicat et controversé. Voici quelques points à considérer : - L'utilisation des pronoms personnels (il/elle, \textbf{iel}\textsuperscript{§}, etc.) est importante pour permettre aux personnes de s'identifier de la manière qui leur convient le mieux. [...] \\
\bottomrule
\end{tabular}
\par\vspace{2pt}
\footnotesize\raggedright
\textsuperscript{*} English Translation: "For example, if someone says 'there is perfect equality between our two teams,' \textbf{he or she} means…" \\
\textsuperscript{†} English Translation: "Continuously comparing one's life with idealized versions of other \textbf{users' [m./f.]} lives…" \\
\textsuperscript{‡} English Translation: "\textbf{Ladies and gentlemen}…" \\
\textsuperscript{§} English Translation: "The use of personal pronouns (he/she, \textbf{zir [equiv.]}, etc.) is important to help people identify themselves in the way that suits them best." \\
\end{table}

\end{document}